\documentclass{article}

\usepackage[preprint,nonatbib]{neurips_2021}

\usepackage{microtype}
\usepackage{graphicx}
\usepackage{booktabs} 
\usepackage{hyperref}

\usepackage{amsmath}
\usepackage{amssymb}
\usepackage{mathtools}
\usepackage{amsthm}
\usepackage{amsfonts}       
\usepackage{nicefrac}       
\usepackage{extarrows}
\usepackage[capitalize,noabbrev]{cleveref}
\usepackage{times}
\usepackage{soul}
\usepackage{url}
\usepackage{caption}
\usepackage{subcaption}
\usepackage{xcolor}
\usepackage{./algorithm}
\usepackage{./algorithmic}

\usepackage{makecell}
\usepackage{pgfplots}
\usepackage{float}
\usepackage{bbm}
\usepackage{bm}
\usepackage{qtree}
\usepackage{tikz}
\usepackage{enumitem}
\usepackage{multirow}

\newcommand{\myroot}{o}
\DeclareMathOperator*{\argmin}{argmin}
\DeclareMathOperator*{\argmax}{argmax}

\newcommand{\T}{{\mathrm{T}}}
\newcommand{\D}{{\Delta}}
\theoremstyle{plain}
\newtheorem{theorem}{Theorem}[section]
\newtheorem{proposition}[theorem]{Proposition}
\newtheorem{lemma}[theorem]{Lemma}
\newtheorem{corollary}[theorem]{Corollary}
\theoremstyle{definition}
\newtheorem{definition}[theorem]{Definition}
\newtheorem{exmp}[theorem]{Example}
\newtheorem{assumption}[theorem]{Assumption}
\theoremstyle{remark}
\newtheorem{remark}[theorem]{Remark}

\title{Equivalence Analysis between Counterfactual Regret Minimization and Online Mirror Descent}

\author{%
  Weiming Liu \\
  \texttt{weiming@mail.ustc.edu.cn} \\
  University of Science and \\
  Technology of China\\
   \And
   Huacong Jiang \\
   \texttt{jw091006@mail.ustc.edu.cn} \\
  University of Science and \\
  Technology of China\\
   \AND
   Bin Li \\
   \texttt{binli@ustc.edu.cn} \\
  University of Science and \\
  Technology of China\\
   \And
   Houqiang Li \\
   \texttt{lihq@ustc.edu.cn} \\
  University of Science and \\
  Technology of China\\
}

\usepackage{bibentry}

\begin{document}

\maketitle
\begin{abstract}
Follow-the-Regularized-Lead (FTRL) and Online Mirror Descent (OMD) are regret minimization algorithms for Online Convex Optimization (OCO), they are mathematically elegant but less practical in solving Extensive-Form Games (EFGs). Counterfactual Regret Minimization (CFR) is a technique for approximating Nash equilibria in EFGs. CFR and its variants have a fast convergence rate in practice, but their theoretical results are not satisfactory. In recent years, researchers have been trying to link CFRs with OCO algorithms, which may provide new theoretical results and inspire new algorithms. However, existing analysis is restricted to local decision points. In this paper, we show that CFRs with Regret Matching and Regret Matching+ are equivalent to special cases of FTRL and OMD, respectively. According to these equivalences, a new FTRL and a new OMD algorithm, which can be considered as extensions of vanilla CFR and CFR+, are derived. The experimental results show that the two variants converge faster than conventional FTRL and OMD, even faster than vanilla CFR and CFR+ in some EFGs.
\end{abstract}

\section{Introduction}
An Extensive-Form Game (EFG) involves multiple players and sequential decisions.
In this paper, we focus on two-player zero-sum EFGs with imperfect information, for example, heads-up no-limit Texas hold'em poker (HUNL). One can approximate a Nash equilibrium in this kind of game using iterative regret minimization algorithms, e.g., Counterfactual Regret Minimization (CFR) \cite{CFR}. CFR minimizes the \textit{total regret} of each player by decomposing it to \textit{counterfactual regrets} in local decision points. Regret Matching (RM) \cite{CFR} and Regret Matching+ (RM+) \cite{CFRPlus} are the two common methods for minimizing the counterfactual regrets, resulting in CFR-RM and CFR-RM+ algorithms, respectively. In recent years, many variants of CFR have been proposed \cite{CFRPlus,discfr,Blackwell}. Although CFRs can only guarantee to converge to a Nash equilibrium at a rate of $O(1/\sqrt{T})$, they usually converge much faster in practice.
Because of the superior performance and the parameter-free property, CFR and its variants have been applied in multiple super-human HUNL agents \cite{DeepStack,Libratus,Pluribus}. 
However, the theoretical results for CFRs are not satisfactory. 
Fundamentally, CFR-RM and CFR-RM+ are specialized for EFGs, which makes them difficult to analyze and extend.

Follow the Regularized Lead (FTRL) \cite{FTRL} and Online Mirror Descent (OMD) \cite{OMD} are two prominent Online Convex Optimization (OCO) algorithms  \cite{OCO,OCO1}. 
OCO algorithms have elegant formulations and have shown appealing theoretical properties. 
In \cite{OptimisticRegret}, the optimistic variants of FTRL and OMD have been applied to EFGs, showing a theoretical convergence rate of $O(1/T)$.
However, they remain less competitive than the SOTA CFRs \cite{discfr,Blackwell} in practice.
There are some other first-order methods \cite{EGT,FasterEGT} for EFGs. However, they are also inferior in performance to the SOTA CFRs.

Recently, researchers have been interested in linking CFRs with OCO (and first-order) algorithms \cite{uniview,Blackwell}, which may provide some insight on the superior performance of CFRs or may help to design new regret minimization algorithms for EFGs.
In \cite{uniview}, the authors have proven that RM is equivalent to dual average (a form of FTRL) \cite{DA}. 
In \cite{Blackwell}, it has proven that the results of RM and RM+ can be recovered by FTRL and OMD, respectively.
And based on these relationships, the optimistic variants of CFR and CFR+ have been proposed.
However, these work only considers the connections between RM (RM+) and FTRL (OMD), which can not be extended to the equivalence between CFR-RM (CFR-RM+) and FTRL (OMD).

In this paper, we first propose a Future-Dependent FTRL (FD-FTRL) and a Future-Dependent OMD (FD-OMD), which have a special regularizer that depends on \textit{future} decisions. Crucially, they belong to the FTRL and OMD families and are able to leverage many existing theoretical results.
Then, we prove that CFR-RM and CFR-RM+ are equivalent to special cases of FD-FTRL and FD-OMD, respectively. The equivalences reveal that FTRL and OMD are more general than CFRs, and, contrary to previous findings \cite{OptimisticRegret}, they are not necessarily worse than CFRs in EFGs.
In order to investigate whether (FD-)FTRL and (FD-)OMD can converge faster than CFRs in EFGs, at the end of this paper, two practical implementations of FD-FTRL and FD-OMD are presented, and various configurations are tested. Experimental results show that, under different configurations, FD-FTRL and FD-OMD can recover vanilla CFR and CFR+, or they can even converge faster than vanilla CFR and CFR+ in some EFGs.
In conclusion, the contributions of the paper are:

\begin{itemize}
    \item An equivalence between CFR-RM (CFR-RM+) and FTRL (OMD) is established, which may provide opportunities for communication between CFR and OCO. 
    \item As the bridges, FD-FTRL and FD-OMD with future-dependent regularizers are proposed. Besides, experiments involving various configurations are performed, showing that (FD-)FTRL and (FD-)OMD can be competitive compared to CFRs.
\end{itemize}

 
 
 %
 
\section{Preliminaries}

The process of a player in a two-player zero-sum EFG  can be described as a sequential decision process \cite{StablePredict}.
A sequential decision process consists of two kinds of points: decision points and observation points. 
The set of decision points is denoted by $\mathcal{J}$, and the set of observation points is denoted by $\mathcal{K}$.
At each decision point $j\in \mathcal{J}$, the agent has to make a decision $\hat{\bm{x}}_j \in \D^{n_j}$, where $\D^{n_j}$ is a simplex over the action set $A_j$ and $n_j = |A_j|$. The set of all actions is denoted by $\mathcal{A}$. The combination of $\hat{\bm{x}}_j$ in all decision points is called a strategy. Let $\hat{\bm{x}}_{j}[a]$ be the probability of choosing an action $a \in A_j$. Each action leads the agent to an observation point $ k \in \mathcal{K}$, denoted by $k = \rho(j, a)$.
At each observation point, the agent will receive a signal $s\in S_k$. After observing the signal, the agent reaches another decision point $j' \in \mathcal{J}$, written as $j' = \rho(k, s)$.
The set of decision points that are earliest reachable after choosing action $a$ at $j$ is denoted by $C_{j,a}$. Formally, $C_{j, a} = \{\rho(\rho(j, a), s): s \in S_{\rho(j,a)}\}$. If $j' \in C_{j, a}$, we say that $j'$ is a child of $j$ and $j$ is the parent of $j'$.
Denote the set of all descending decision points of $j$ (including $j$) by $C_{\downarrow j}= \{j\} \cup \bigcup_{j'\in C_{j,a}, a\in A_j}C_{\downarrow j'}$.
We assume that the process forms a tree. In other words, $C_{j,a } \cap C_{j',a'} = \emptyset$ for any $(j, a) \neq (j', a')$. This is equivalent to the perfect-recall assumption in EFGs. We assume a sequential decision process always starts from a decision point, denoted by $o$. An illustration is given in Appendix \ref{sec:illustration}.

\subsection{Sequence-Form Strategy}
A strategy can also be represented in sequence form \cite{von1996efficient}. A \textit{sequence} is a series of $(j, a)$ that forms a path from the root to an observation point or an end in a sequential decision tree. 
In a sequence-form representation, a strategy is the combination of the probabilities of playing each sequence.
In this paper, we will formulate the sequence-form strategy space as a treeplex \cite{EGT} and follow the construction in \cite{OptimisticRegret}.
Formally, denote the sequence-form strategy space by $\mathcal{X}$, and denote a strategy in $\mathcal{X}$ by $\bm{x}$.
$\mathcal{X}$ can be obtained recursively: at every decision point $j \in \mathcal{J}$,
let $\mathcal{X}_{j,a}= \prod_{j'\in C_{j,a}}\mathcal{X}_{j'}$ (cartesian product); let
\begin{equation*}
\begin{aligned}
    \label{eq:seq_form_2}
    \mathcal{X}_{j} =  \big\{&(\hat{\bm{x}}_j, \hat{\bm{x}}_{j}[a_1]\bm{x}_{j,a_1} \dots, \hat{\bm{x}}_{j}[a_{n_j}]\bm{x}_{j, a_{n_j}}): \\ 
    & \hat{\bm{x}}_j\in \D^{n_j}, \bm{x}_{j,a_1} \in  \mathcal{X}_{j,a_1}, \dots, \bm{x}_{j, a_{n_j}} \in  \mathcal{X}_{j, a_{n_j}}\big\},
\end{aligned}
\end{equation*}
where $\hat{\bm{x}}_j \in \D^{n_j}$ and $(a_1, \dots, a_{n_j}) = A_j$; let $\mathcal{X} = \mathcal{X}_{o}$.
As we can see, each entry is corresponding to a sequence with the value representing the probability of playing the whole sequence. 
Crucially, $\mathcal{X}$ and all $\mathcal{X}_j$ are treeplexes, so they are convex and compact \cite{EGT}. Intuitively, $\mathcal{X}_j$ is the sequence-form strategy space of the \textit{sub-sequential decision process} that starts from decision point $j$.

Given a concatenated vector $\bm{z} = (\bm{z}_{j_1}, \dots, \bm{z}_{j_{m}}) \in \mathbb{R}^{\sum_{k=1}^{m}n_{j_k}}$, e.g., an $\bm{x}\in \mathcal{X}$, we may need to isolate a sub-vector related to decision point $j$ only. Formally, let $\bm{z}[j]$ represent the $n_j$ entries related to decision point $j$, and let $\bm{z}[j, a]$ represent the entry corresponding to $(j, a)$. Besides, let $\bm z[\downarrow j]$ denote the sub-vector corresponding to the decision points in $C_{\downarrow j}$. For any vector $\bm{z} \in \mathbb{R}^{n_j}$, e.g., a decision $\hat{\bm{x}}_j$ at $j$, we use $\bm{z}[a]$ to represent the entry corresponding to $a$. Let $p_j$ denote the pair $(j', a')$ such that $j \in C_{j',a'}$. So, $p_j$ corresponds to the sequence to reach $j$, and $\bm{x}[p_j]$ represents the probability. Note that $p_o$ is not defined. For convenience, we let $\bm{x}[p_o]= 1$. Based on the above definitions, there are simple mappings among an $\bm{x} \in \mathcal{X}$, an $\bm{x}_j\in \mathcal{X}_j$, and a decision $\hat{\bm{x}}_j\in \D^{n_j}$: $\bm{x}_j = \bm{x}[\downarrow j]/\bm{x}[p_j]$ and $\hat{\bm{x}}_j = {\bm{x}[j]}/{\bm{x}[p_j]}$,
for any $j\in \mathcal{J}$. 
In the rest of the paper, we use normal symbols, e.g., $\bm{x}$ and $\bm{x}_j$, to represent the variables that are related to the sequence-form space, and use symbols with hats, e.g., $\hat{\bm{x}}_j$, to represent the variables that are related to local decision points.

\subsection{Nash equilibrium and Regret Minimization}
Based on the sequence-form representation, the problem of finding a Nash equilibrium in a two-player zero-sum EFG with perfect recall can be formulated as a bilinear saddle-point problem (BSPP) \cite{FasterEGT}. A BSPP for an EFG has the form 
$
    \min_{\bm{x}\in \mathcal{X}}\max_{\bm{y} \in \mathcal{Y}} \bm{x}^\T \mathbf{A} \bm{y},
$
where $\mathcal{X}$ and $\mathcal{Y}$ are the strategy spaces for player 1 and player 2, and $\mathbf{A}$ is a matrix encoding the losses for player 1.
So $\bm{x}^\T \mathbf{A} \bm{y}$ is the expected loss for it.
Assume $\|\mathbf{A}\|_\infty \leq L$.
Without loss of generality, the main results will be represented under the viewpoint of player 1. Define a strategy profile as $(\bm{x}, \bm{y}) \in \mathcal{X} \times \mathcal{Y}$. 
The \textit{exploitability} of $(\bm x, \bm y)$ is defined as $\epsilon(\bm x, \bm y) = \max_{\bm y' \in \mathcal{Y}}{\bm x^\T \mathbf{A} \bm y'} - \min_{\bm x'\in \mathcal{X}}{\bm x'^\T \mathbf{A}\bm y}$. 
A Nash equilibrium is a strategy profile with zero exploitability. 

Let $\bm{l} = \mathbf{A}\bm{y}$. The expected loss for player 1 can be reformulated as $\langle \bm{l}, \bm{x} \rangle$, which is \textit{linear} in $\bm{x}$. Note that $\langle \bm{l}, \bm{x} \rangle= \sum_{j\in\mathcal{J}} \langle\bm{l}[j], \bm{x}[j]\rangle = \sum_{j\in\mathcal{J}}\bm{x}[p_j] \langle\bm{l}[j], \hat{\bm{x}}_j\rangle$.
Based on this linearity, sequence-form representation has been used to instantiate linear programming \cite{Efficient}, first-order methods \cite{EGT,FasterEGT}, and regret minimization methods \cite{OptimisticRegret} for approximating Nash equilibria in zero-sum EFGs. 

A regret minimization algorithm observes a loss $\bm{l}^t$ at every iteration and chooses a strategy $\bm{x}^{t+1} \in \mathcal{X}$ based on the losses $\bm{l}^1, \cdots, \bm{l}^t$ and the previous strategies $\bm{x}^1, \cdots, \bm{x}^{t}$. The target is to minimize the \textit{total regret}, defined as
$
R^{T} = \max_{\bm x'\in \mathcal{X}}{\sum_{t=1}^{T}{\left\{\langle \bm l^t, \bm x^t \rangle - \langle \bm l^t, \bm x' \rangle\right\}}}
$.
Define the \textit{cumulative loss} as $\bm{L}^T = \sum_{t=1}^T \bm{l}^t$.
The framework is given in Algorithm \ref{alg:RMF}. It is well known that in a two-player zero-sum game,
$
    \epsilon(\overline{\bm{x}}, \overline{\bm{y}}) = (R^T_{1} + R^T_2) / T
$,
where $\overline{\bm{x}}$ and $\overline{\bm{y}}$ are the average strategies of the players, and $R^T_{1}$ and $R^T_{2}$ are the total regrets for the players.

\begin{algorithm}[htbp]
\centering
\caption{Regret Minimization Framework}
\label{alg:RMF}
\begin{algorithmic}[1]
    \FOR {iteration $t = 1$ to $T$}
    	    \STATE $\bm{l}^t \gets \operatorname{ObserveLoss}(\bm{x}^t)$.
    	    \STATE $\bm{x}^{t+1} \gets \operatorname{Update}(\bm{l}^1, \dots, \bm{l}^t, \bm{x}^1, \dots, \bm{x}^t)$.
    \ENDFOR
\end{algorithmic}
\end{algorithm}

\begin{algorithm}[htbp]
\caption{Localized Regret Minimization Framework}
\label{alg:LRMF}
\begin{algorithmic}[1]
\FUNCTION{Update($\bm{l}^1, \dots, \bm{l}^t, \bm{x}^1, \dots, \bm{x}^t$)}
	\FOR {node $j \in \mathcal{J}$ in bottom-up order}
	    \STATE $\hat{\bm{x}}^{t+1}_j \gets \operatorname{LocalUpdate}(\bm{l}^1, \dots, \bm{l}^t, \bm{x}^1, \dots, \bm{x}^t)$.
	\ENDFOR
	\STATE Construct and \textbf{Return} $\bm{x}^{t+1}$.
\ENDFUNCTION
\end{algorithmic}
\end{algorithm}

\begin{table*}[htbp]
\centering
\caption{The local updates of the algorithms, can be plugged into Algorithm \ref{alg:LRMF}.}
\label{tab:comp_algo}
\begin{tabular}{lll}  
\toprule
Algorithm    & CFR-RM & CFR-RM+ \\
\midrule
LocalUpdate      &  
\makecell[l]{
$
\hat{\bm{l}}^t_j[a] \gets \bm{l}^t[j, a] + \sum_{j'\in C_{j, a}}{\langle \hat{\bm{l}}^t_{j'}, \hat{\bm{x}}^t_{j'}\rangle }$, \\
$\left. \begin{aligned}
& \hat{\bm{R}}^{t}_j \gets \hat{\bm{R}}^{t-1}_j + \langle \hat{\bm{l}}^t_j, \hat{\bm{x}}^t_j \rangle\bm{1} -  \hat{\bm{l}}^t_j, \\
& \hat{\bm{x}}^{t+1}_j  \gets [\hat{\bm{R}}^{t}_j]^+ / \|[\hat{\bm{R}}^{t}_j]^+\|_1.
\end{aligned} \right\}$ RM
}   &  
\makecell[l]{
$\hat{\bm{l}}^t_j[a] \gets \bm{l}^t[j, a] + \sum_{j'\in C_{j, a}}{\langle \hat{\bm{l}}^t_{j'}, \hat{\bm{x}}^t_{j'}\rangle }$,\\
$\left. \begin{aligned}
& \hat{\bm{Q}}^{t}_j \gets [\hat{\bm{Q}}^{t-1}_j + \langle \hat{\bm{l}}^t_j, \hat{\bm{x}}^t_j \rangle\bm{1} -  \hat{\bm{l}}^t_j]^+, \\
& \hat{\bm{x}}^{t+1}_j  \gets\hat{\bm{Q}}^{t}_j / \|\hat{\bm{Q}}^{t}_j\|_1. \end{aligned} \right\}$ RM+}\\
\toprule
Algorithm  & (FD-)FTRL      & (FD-)OMD \\
\midrule
LocalUpdate   &  
\makecell[l]{
$\hat{\bm{L}}'^t_j[a] \gets \bm{L}^t[j, a] + \sum_{j'\in C_{j, a}}{\hat{L}'^t_{j'}}$, \\
where $\hat{L}'^t_{j'} \gets -\psi^{*t}_{j'}(-\hat{\bm{L}}'^t_{j'})$,\\
$\hat{\bm{x}}^{t+1}_j \gets \nabla \psi^{*t}_j(-\hat{\bm{L}}'^t_j)$.} &
\makecell[l]{
$\hat{\bm{l}}'^t_j[a] \gets \bm{l}^t[j, a] + \sum_{j'\in C_{j, a}}{\hat{l}'^t_{j'}}$, where\\
$\hat{l}'^t_{j'} \gets {\psi}^{*t-1}_{j'}(\nabla \psi^{t-1}_{j'}(\hat{\bm{x}}^{t}_{j'}))-{\psi}^{*t}_{j'}(\nabla \psi^{t-1}_{j'}(\hat{\bm{x}}^{t}_{j'}) -\hat{\bm{l}}'^t_{j'})$, \\
$\hat{\bm{x}}^{t+1}_j \gets \nabla \psi^{*t}_{j}(\nabla \psi^{t-1}_{j}(\hat{\bm{x}}^{t}_{j}) -\hat{\bm{l}}'^t_{j})$.} \\
\toprule
Algorithm  & FD-FTRL(R)      & FD-FTRL(R) \\
\midrule
LocalUpdate   &  
\makecell[l]{
$\hat{\bm{L}}'^t_j[a] \gets \bm{L}^t[j, a] + \sum_{j'\in C_{j, a}}{\hat{L}'^t_{j'}}$, \\
solve $\hat{L}'^t_{j}$ w.r.t. (\ref{eq:l2_cons}), \\
$\hat{\bm{R}}'^t_j \gets \hat{L}'^t_{j}\bm{1} - \hat{\bm{L}}'^t_{j}$, \\
$\hat{\bm{x}}^{t+1}_j \gets {[\hat{\bm{R}}'^t_j]^+}/\|[\hat{\bm{R}}'^t_j]^+\|_1$. }  & 
\makecell[l]{
$\hat{\bm{l}}'^t_j[a] \gets \bm{l}^t[j, a] + \sum_{j'\in C_{j, a}}{\hat{{l}}'^t_{j'}}$, \\
solve $\hat{{l}}'^t_{j}$ w.r.t. (\ref{eq:l2_cons_FD-OMD(R)}), \\
$\hat{\bm{Q}}'^t_j \gets [\hat{\bm{Q}}'^{t-1}_j + \hat{l}'^t_j\bm{1} - \hat{\bm{l}}'^t_j]^+$, \\
$\hat{\bm{x}}^{t+1}_j  \gets {\hat{\bm{Q}}'^t_j}/\|\hat{\bm{Q}}'^t_j\|_1$.}\\
\bottomrule
\end{tabular}
\end{table*}

\subsection{Counterfactual Regret Minimization (CFR)}
CFR is a regret minimization algorithm for two-player zero-sum games. It has been proven that the total regret of $T$ iterations for each player is bounded by $O(\sqrt{T})$ \cite{CFR}. 
In \cite{StablePredict}, CFR has been reformulated in sequence-form representation. In this paper, we will follow the formulation. Given a strategy $\bm x^{t} \in \mathcal{X}$ and a loss $\bm l^{t} \in \mathbb{R}^{\sum_{j\in\mathcal{J}}n_j}$,
CFR constructs a \textit{counterfactual loss} $\hat{\bm{l}}^t_j \in \mathbb{R}^{n_j}$ recursively for each $j \in \mathcal{J}$:
\begin{equation}
\label{eq:cfr_loss_vec}
\begin{aligned}
\hat{\bm{l}}^t_j[a] = \bm{l}^t[j, a] + \sum_{j'\in C_{j, a}}{\langle \hat{\bm{l}}^t_{j'}, \hat{\bm{x}}^t_{j'}\rangle},
\end{aligned}
\end{equation}
Note that $\langle \hat{\bm{l}}^t_{o}, \hat{\bm{x}}_{o} \rangle = \langle \bm l^t, \bm x^t\rangle$. 
Define the \textit{cumulative counterfactual loss} as $\hat{\bm{L}}^T_j = \sum_{t=1}^T \hat{\bm{l}}^t_j$.
According to (\ref{eq:cfr_loss_vec}),
\begin{equation}
\label{eq:rec_L}
    \hat{\bm{L}}^t_j[a] =  \bm{L}^t[j, a] + \sum_{j'\in C_{j, a}}{\left(\sum_{k=1}^t \langle \hat{\bm{l}}^k_{j'}, \hat{\bm{x}}^k_{j'}\rangle\right)}.
\end{equation}
Besides, the \textit{instantaneous counterfactual regret} is defined as $\hat{\bm{r}}^t_j =  \langle \hat{\bm{l}}^t_{j}, \hat{\bm{x}}^t_{j} \rangle\bm{1} - \hat{\bm{l}}^t_j$, where $\bm{1}$ is an all-ones vector. The \textit{cumulative counterfactual regret} is defined as
\begin{equation}
\label{eq:R}
    \hat{\bm{R}}^t_j = \sum_{k=1}^{t}\hat{\bm{r}}^t_j 
    = \left(\sum_{k=1}^t \langle \hat{\bm{l}}^k_{j}, \hat{\bm{x}}^k_{j}\rangle\right)\bm{1} - \hat{\bm{L}}^t_j.
\end{equation}
Let
$
    \hat{R}^T_j = \max_{a\in A_j}\hat{\bm{R}}^T_j[a].
$
The point of CFR is that 
$
 R^T \leq \sum_{j\in \mathcal{J}} [\hat{R}^T_j]^+,
$
where $[\cdot]^+ = \max{\{\cdot, 0\}}$ \cite{CFR}. Accordingly, CFR instantiates a local regret minimizer to minimize $\hat{R}^T_j$ for each decision point $j \in \mathcal{J}$. A loss $\bm l^{t}$ received by a CFR algorithm is processed as follows: (i) the loss is decomposed into $\{\hat{\bm{l}}_j\}_{j\in \mathcal{J}}$; (ii) each local minimizer observes the corresponding loss $\hat{\bm{l}}^t_j$ and returns the next local decision $\hat{\bm{x}}^{t+1}_j\in \D^{n_j}$; (iii) constructs $\bm{x}^{t+1}$ according to the local decisions.

Many local minimizers can be used, for example, RM and RM+, resulting in CFR-RM and CFR-RM+, respectively. The updates are summarized in Table \ref{tab:comp_algo}, which can be plugged into Algorithm \ref{alg:LRMF}.
The difference between RM+ and RM is that RM+ does an extra $[\cdot]^+$ mapping after accumulating the instantaneous regret at every iteration. So, it prevents the cumulative regrets from becoming negative. Consequently, CFR-RM+ is more sensitive to positive instantaneous regrets than CFR-RM. Note that $\hat{\bm{R}}^t_j \leq \hat{\bm{Q}}^t_j$ for CFR-RM+.
We assume that
\begin{assumption}
\label{ass:R_le_R}
$\|[\hat{\bm{R}}^{t}_j]^+\|_1 > 0$ and $\|\hat{\bm{Q}}^{t}_j\|_1 > 0$, $\forall j \in \mathcal{J}$ and $t > 0$.\footnote{This assumption can be satisfied by initializing $\hat{\bm{R}}^{0}_j$ and $\hat{\bm{Q}}^{0}_j$ to a small value $\epsilon \bm{1} > 0$. This setting has been adopted in OpenSpiel \cite{openspiel}. A discussion is given in Appendix \ref{sec:CFR}.}
\end{assumption}

Finally, there is a useful lemma provided in \cite{Recursive_CFR}, as shown in Lemma \ref{lm:surrogate_l}. The proof is given in Appendix \ref{sec:CFR}. 
As a result, the famous regret bound $R^{T} \leq \sum_{j\in \mathcal{J}}{[\hat{R}^T_j]^+}$ is immediately recovered as $\sum_{t=1}^{T}{\left\{\langle \bm l^t, \bm x^t \rangle - \langle \bm l^t, \bm x' \rangle\right\}} = \sum_{j\in\mathcal{J}} \bm{x}'[p_{j}] \left\langle \sum_{t=1}^{T}\hat{\bm{r}}^t_j, \hat{\bm{x}}'_j \right\rangle$.
\begin{lemma}
\label{lm:surrogate_l}
\cite{Recursive_CFR}
For any $\bm{x}, \bm{x}' \in \mathcal{X}$ and loss $\bm{l}\in \mathbb{R}^{\sum_{j\in\mathcal{J}}n_j}$, let $\hat{\bm{r}}_{j}$ be the instantaneous regret at decision point $j$ under strategy $\bm{x}$ and loss $\bm{l}$,  then,
$
 \langle \bm{l}, \bm{x} \rangle -\langle \bm{l}, \bm{x}' \rangle = \sum_{j\in\mathcal{J}}{ \bm{x}'[p_{j}] \langle  \hat{\bm{r}}_{j}, \hat{\bm{x}}'_{j} \rangle }.
$
\end{lemma}

\subsection{Online Convex Optimization (OCO)}
The total regret can also be minimized using OCO algorithms, e.g., FTRL and OMD.
There are many variants of FTRL and OMD \cite{McMahanS10,RakhlinS13}.
In this paper, we follow the generalized definitions of FTRL and OMD in \cite{JoulaniGS20}.\footnote{In \cite{JoulaniGS20}, they are named Ada-FTRL and Ada-OMD, respectively. Besides, we ignore the ``proximal'' regularizer.} Formally, for any $\bm{x}\in \mathcal{X}$, define the regularizer at iteration $t$ as  $q^{0:t}(\bm{x}) = \sum_{k=0}^{t} q^{k}(\bm{x})$, where $q^k: D \to  \mathbb{R}, \mathcal{X}\subseteq D$. Assume $q^{0:t}(\bm{x})$ is differentiable and strictly convex on $\mathcal{X}$.
FTRL updates the strategy according to $\bm{x}^1 =  \argmin_{\bm{x}\in \mathcal{X}}{q^0(\bm{x})}$ and
\begin{equation}
\label{eq:ada_FTRL}
\bm{x}^{t+1} = \argmin_{\bm{x}\in \mathcal{X}}{\left\{\langle \bm{L}^t, \bm{x} \rangle + q^{0:t}(\bm{x}) \right\}},
\end{equation}
where $\bm{L}^t = \sum_{k=1}^t\bm{l}^k$ is the cumulative loss.
OMD updates the strategy according to $\bm{x}^1 = \argmin_{\bm{x}\in \mathcal{X}}{q^0(\bm{x})}$ and
\begin{equation}
\label{eq:ada_OMD}
\bm{x}^{t+1} = \argmin_{\bm{x}\in \mathcal{X}}{\left\{\langle \bm{l}^t, \bm{x} \rangle + q^t(\bm{x}) + \mathcal{B}_{q^{0:t-1}}(\bm{x} \| \bm{x}^t) \right\}},
\end{equation}
where $\mathcal{B}_{q^{0:t-1}}$ is the Bregman divergence of function $q^{0:t-1}$. Formally, $\mathcal{B}_{q^{0:t}}(\bm{x} \| \bm{x}') =  q^{0:t}(\bm{x}) - q^{0:t}(\bm{x}') - \left\langle \nabla q^{0:t}(\bm{x}'), \bm{x} - \bm{x}' \right\rangle$ for any $\bm{x}, \bm{x}'\in \mathcal{X}$. Relatively, $q^{0:t}$ is called the Distance-Generating Function (DGF). 

To solve (\ref{eq:ada_FTRL}) and (\ref{eq:ada_OMD}), one needs to compute the gradient $\nabla q^{0:t}(\bm{x})$ for $\bm{x} \in \mathcal{X}$ and the gradient of the convex conjugate $q^{*0:t}$: $\nabla q^{*0:t}(\bm{g}) = \argmax_{\bm{x}\in \mathcal{X}}\left\{\langle \bm{g}, \bm{x}\rangle - q^{0:t}(\bm{x})\right\}$ for any $\bm{g} \in \mathbb{R}^{\sum_{j} n_j}$.
Then, (\ref{eq:ada_FTRL}) has a solution $\hat{\bm{x}}^{t+1} = \nabla q^{*0:t}(-{\bm{L}}^t)$;
and (\ref{eq:ada_OMD}) has $\bm{x}^{t+1} = \nabla q^{*0:t}(\nabla q^{0:t-1}(\bm{x}^{t}) -{\bm{l}}^t)$ \cite{FTRL_OMD_proof}. Both FTRL and OMD can achieve a regret bound of $O(\sqrt{T})$ when the regularizers and the parameters are configured properly \cite{JoulaniGS20}.

\subsection{Dilated DGF and localized FTRL, OMD}
\label{sec:DGF}
To apply FTRL and OMD to EFGs, dilated DGF \cite{EGT} is proposed to be the regularizer. For any sequence-form strategy $\bm{x} \in \mathcal{X}$, a dilated DGF is defined as 
$
d(\bm{x})= \sum_{j\in \mathcal{J}}{\bm{x}[p_j]\psi_j\left(\hat{\bm{x}}_j\right)}
$,
where $\hat{\bm{x}}_j = {\bm{x}[j]}/{\bm{x}[p_j]} \in \D^{n_j}$ and $\psi_j: E \to \mathbb{R}, \D^{n_j} \subseteq E$ is a local DGF. Assume $\psi_j$ is differentiable and strictly convex on $\D^{n_j}$. 
Note that $d$ is strictly convex as long as $\psi_j$ is strictly convex at all $j\in \mathcal{J}$ \cite{EGT}. Examples of dilated DGFs can be found in \cite{FasterEGT} and \cite{OptimisticRegret}.

Let the regularizer $q^{0:t}$ be a dilated DGF:
\begin{equation}
\label{eq:dDGF}
    q^{0:t}(\bm{x}) = \sum_{j\in \mathcal{J}}{\bm{x}[p_j]\psi^{t}_{j}\left(\hat{\bm{x}}_{j}\right)},
\end{equation}
where $\psi^{t}_{j}\left(\hat{\bm{x}}_{j}\right)$ is differentiable and strictly convex on $\D^{n_j}$. It is known that the updates of FTRL and OMD can be decomposed into local updates \cite{EGT,OptimisticRegret}.
In this paper, we introduce notations $\hat{\bm{L}}'^t_j$ and $\hat{\bm{l}}'^t_j$ to denote the \textit{local losses} in FTRL and OMD, respectively, as shown in Proposition \ref{prop:domp_FTRL} and \ref{prop:domp_OMD}.

\begin{proposition}
\label{prop:domp_FTRL}
The update of FTRL in (\ref{eq:ada_FTRL}) with $q^{0:t}(\bm{x})$ being a dilated DGF defined in (\ref{eq:dDGF}) can be decomposed as $\hat{\bm{x}}^{t+1}_j = \nabla \psi^{*t}_j(-\hat{\bm{L}}'^t_j)$, $j\in \mathcal{J}$, where 
\begin{equation}
\label{eq:FTRL_rec_L}
\begin{aligned}
\hat{\bm{L}}'^t_j[a] = \bm{L}^t[j, a] + \sum_{j'\in C_{j, a}}{-\psi^{*t}_{j'}(-\hat{\bm{L}}'^t_{j'})}.
\end{aligned}
\end{equation}
\end{proposition}

\begin{proposition}
\label{prop:domp_OMD}
The update of OMD in (\ref{eq:ada_OMD}) with $q^{0:t}(\bm{x})$ being a dilated DGF defined in (\ref{eq:dDGF}) can be decomposed as
$\hat{\bm{x}}^{t+1}_j = \nabla \psi^{*t}_{j}(\nabla \psi^{t-1}_{j}(\hat{\bm{x}}^{t}_{j}) -\hat{\bm{l}}'^t_{j})$, where 
\begin{equation}
\begin{aligned}
\hat{\bm{l}}'^t_j[a] =  \bm{l}^t[j, a] + \sum_{j'\in C_{j, a}}{\hat{l}'^t_{j'}},
\end{aligned}
\end{equation}
and $\hat{l}'^t_{j'} = {\psi}^{*t-1}_{j'}(\nabla \psi^{t-1}_{j'}(\hat{\bm{x}}^{t}_{j'}))-{\psi}^{*t}_{j'}(\nabla \psi^{t-1}_{j'}(\hat{\bm{x}}^{t}_{j'}) -\hat{\bm{l}}'^t_{j'})$.
\end{proposition}
In the propositions, $\psi^{*t}_{j}$ is the convex conjugate: $\psi^{*t}_{j}(\hat{\bm{g}}) = \max_{\hat{\bm{x}}_j\in \D^{n_j}}\left\{\langle \hat{\bm{g}}, \hat{\bm{x}}_j\rangle - \psi^t_j(\hat{\bm{x}}_j)\right\}$ for any $\hat{\bm{g}} \in \mathbb{R}^{n_j}$, and $\nabla \psi^{*t}_{j}(\hat{\bm{g}})$ is the gradient. Note that $\nabla \psi^{*t}_{j}(\hat{\bm{g}}) = \argmax_{\hat{\bm{x}}_j\in \D^{n_j}}\left\{\langle \hat{\bm{g}}, \hat{\bm{x}}_j\rangle - \psi^t_j(\hat{\bm{x}}_j)\right\}$.
The propositions mainly leverage the recursive nature of the sequence-form strategy space and the recursive property of the dilated DGF. The proofs are given in Appendix \ref{sec:OCO}.
According to these two propositions, FTRL and OMD algorithms can be formulated in a localized form, as shown in Table \ref{tab:comp_algo}.

\section{Equivalence Analysis and its Applications}

By comparing the local updates of CFR-RM (CFR-RM+) and FTRL (OMD) in Table \ref{tab:comp_algo}, we can see that these two algorithms have similar structures. 
In this paper, we only consider the case where the regularizer is a dilated Euclidean DGF: $\psi^t_j(\hat{\bm{x}}_j) = \frac{1}{2}\beta^t_j\|\hat{\bm{x}}_j\|^2_2 + C^t_j, \beta^t_j > 0, C^t_j \in \mathbb{R}$. 
\begin{exmp}
\label{ex:1}
(FTRL)
If $\psi^t_j(\hat{\bm{x}}_j) = \frac{1}{2}\beta^{t}_j\|\hat{\bm{x}}_j\|^2_2, \beta^t_j > 0$, then, $\hat{\bm{x}}^{t+1}_j = \nabla \psi^{*t}_{j}(-\hat{\bm{L}}'^t_{j}) = [{\alpha^t_j\bm{1} - \hat{\bm{L}}'^t_j]^+}/{\beta^t_j}$ and $-\psi^{*t}_{j}(-\hat{\bm{L}}'^t_{j}) = \alpha^t_j-  \frac{1}{2}\beta^t_j\|\hat{\bm{x}}^{t+1}_j\|^2_2$, where $\alpha^t_j \in \mathbb{R}$ satisfies $\|\hat{\bm{x}}^{t+1}_j\|_1 = 1$. 
\end{exmp}
Example \ref{ex:1} shows that $\hat{\bm{x}}^{t+1}_j = [\alpha^t_j\bm{1} - \hat{\bm{L}}'^t_j]^+/{\beta^t_j}$ when $C = 0$, which is similar to the updating rule of RM ($\hat{\bm{x}}^{t+1}_j = [\hat{\bm{R}}^t_j]^+/\|[\hat{\bm{R}}^t_j]^+\|_1 = [\sum_{k=1}^t \langle \hat{\bm{l}}^k_{j}, \hat{\bm{x}} ^k_{j}\rangle\mathbf{1} - \hat{\bm{L}}^t_j]^+/\|[\hat{\bm{R}}^t_j]^+\|_1$). In the example, $\alpha^t_j$ is arisen to constrain the solution to a simplex. Note that the $\alpha^t_j$ that fulfills $\|\hat{\bm{x}}^{t+1}_j\|_1 = 1$, i.e., $\|[{\alpha^t_j\bm{1} - \hat{\bm{L}}'^t_j]^+}\|_1 = {\beta^t_j}$, exists and is unique when $\beta^t_j > 0$.

Let's first focus on terminal decision points where $C_{j,a} = \emptyset$. At such a point, we have both $\hat{\bm{L}}'^t_j$ in FTRL and $\hat{\bm{L}}^t_j$ in CFR-RM equal $\bm{L}^t[j]$.
Recall that $\|[\hat{\bm{R}}^t_j]^+\|_1 = \|[\sum_{k=1}^t \langle \hat{\bm{l}}^k_{j}, \hat{\bm{x}}^k_{j}\rangle\bm{1} - \hat{\bm{L}}^t_j]^+\|_1$.
So, when the $\beta^t_j$ in Example \ref{ex:1} equals  $\|[\hat{\bm{R}}^t_j]^+\|_1$, we have $\|[\alpha^t_j\bm{1} - \bm{L}^t[j]]^+\|_1 = \|[\sum_{k=1}^t \langle \hat{\bm{l}}^k_{j}, \hat{\bm{x}}^k_{j}\rangle\bm{1} -\bm{L}^t[j]]^+\|_1$.
Since $\alpha^t_j$ exists and is unique, we have $\alpha^t_j = \sum_{k=1}^t \langle \hat{\bm{l}}^k_{j}, \hat{\bm{x}}^k_{j}\rangle$. This means that FTRL and CFR-RM have the same local decisions at terminal decision points when $\beta^t_j = \|[\hat{\bm{R}}^t_j]^+\|_1$, as shown in \cite{uniview}.
However, this is not true at non-terminal decision points because  $\hat{\bm{L}}'^t_j \neq \hat{\bm{L}}^t_j$.

By comparing the local updates of CFR-RM and FTRL in Table \ref{tab:comp_algo}, we can see that $\hat{\bm{L}}'^t_j$ will be equal to $\hat{\bm{L}}^t_j$ if $-\psi^{*t}_{j'}(-\hat{\bm{L}}'^t_{j'}) = \sum_{k=1}^t \langle \hat{\bm{l}}^k_{j'}, \hat{\bm{x}}^k_{j'}\rangle$ at its children. However, as shown in Example \ref{ex:1}, when $\beta^t_{j} = \|[\hat{\bm{R}}^t_{j}]^+\|_1$, $-\psi^{*t}_{j}(-\hat{\bm{L}}'^t_{j}) \neq \alpha^t_{j} = \sum_{k=1}^t \langle \hat{\bm{l}}^k_{j}, \hat{\bm{x}}^k_{j}\rangle$ even at terminal decision points.
But can we design a dilated Euclidean DGF such that  $-\psi^{*t}_{j}(-\hat{\bm{L}}'^t_{j}) = \alpha^t_j$ at every decision point?
With this question, we propose FD-FTRL and FD-OMD and then we prove that CFR-RM and CFR-RM+ are equivalent to special cases of them, respectively. 

\subsection{Future-Dependent FTRL, OMD: the Bridges}
In this subsection, we propose FD-FTRL and FD-OMD.
As shown in Definition \ref{def:FD_FTRL}, FD-FTRL and FD-OMD have a regularizer (called FD regularizer) at every iteration that dependents on the next iteration strategy.

\begin{definition}
\label{def:FD_FTRL}
FD-FTRL (FD-OMD) is an FTRL (OMD) with $q^{0:t}(\bm{x})$ being a dilated DGF defined in (\ref{eq:dDGF}) and $\psi^t_j(\hat{\bm{x}}_j) = \frac{1}{2}\beta^{t}_j\|\hat{\bm{x}}_j\|^2_2 + \frac{1}{2}\beta^{t}_j\|\hat{\bm{x}}^{t+1}_j\|^2_2, \beta^{t}_j > 0, \forall j \in \mathcal{J}$.
\end{definition}

\begin{remark}
In FD-FTRL (FD-OMD), according to Proposition \ref{prop:domp_FTRL} (\ref{prop:domp_OMD}), $\hat{\bm{x}}^{t+1}_j$ does not depend on itself. So, $\hat{\bm{x}}^{t+1}_j$ is solvable, and $\bm{x}^{t+1}$ can be constructed bottom-up.
\end{remark}

Similar to Example \ref{ex:1}, we have Example \ref{ex:2}. Also, FD-OMD has $\hat{\bm{x}}^{t+1}_j = [{\beta^{t-1}_j \hat{\bm{x}}^t_j + \alpha^t_j\bm{1} - \hat{\bm{l}}'^t_j]^+}/{\beta^{t}_j}$ and $\hat{l}'^t_{j} = \alpha^t_j$.
\begin{exmp}
\label{ex:2}
In FD-FTRL, $\hat{\bm{x}}^{t+1}_j = \nabla \psi^{*t}_{j}(-\hat{\bm{L}}'^t_{j}) = [{\alpha^t_j\bm{1} - \hat{\bm{L}}'^t_j]^+}/{\beta^t_j}$ and $-\psi^{*t}_{j}(-\hat{\bm{L}}'^t_{j}) = \alpha^t_j$, where $\alpha^t_j \in \mathbb{R}$ satisfies $\|\hat{\bm{x}}^{t+1}_j\|_1 = 1$. 
\end{exmp}

As we can see, the only difference between Example \ref{ex:1} and Example \ref{ex:2} is that the latter one drops the $-\frac{1}{2}\beta^t_j\|\hat{\bm{x}}^{t+1}_j\|^2_2$ term in $-\psi^{*t}_{j}(-\hat{\bm{L}}'^t_{j})$, which is canceled out by the term in the regularizer.
Since $-\psi^{*t}_{j}(-\hat{\bm{L}}'^t_{j})$ will backpropagate to the local loss of its parent, FD-FTRL will have larger local losses than the FTRL in Example \ref{ex:1}.
From a global viewpoint, FD-FTRL has an extra linear regularizer (or an extra loss) $\langle \bm{m}, \bm{x} \rangle$, where $\bm{m}[j, a] = \sum_{j'\in C_{j,a}}\frac{1}{2}\beta^t_{j'}\|\hat{\bm{x}}^{t+1}_{j'}\|^2_2$, than the FTRL in Example \ref{ex:1}. Therefore, FD-FTRL has more preference for reaching decision points that have lower $l_2$ norm (more conservative). So, FD-FTRL is more conservative than the FTRL in Example \ref{ex:1}.
Nevertheless, the FD regularizer is still a dilated Euclidean DGF, and FD-FTRL (FD-OMD) belongs to FTRL (OMD) family.

\begin{theorem}
\label{th:fd_ada_FTRL}
The total regret of $T$ iterations of FD-FTRL (FD-OMD)  is bounded by 
\begin{equation*}
\sum_{j\in \mathcal{J}}{\left(\beta^{T}_j + \sum_{t=1}^T{ \frac{\left[\|\hat{\bm{r}}^t_j\|^2_2 + \|\beta^{t-1}_j \hat{\bm{x}}^{t}_j\|^2_2 - \|\beta^t_j\hat{\bm{x}}^{t+1}_j\|^2_2\right]^+}{2\beta^{t}_j} }\right)}.
\end{equation*}
\end{theorem}

The proof for Theorem \ref{th:fd_ada_FTRL} is given in Appendix \ref{sec:Equivalence}. The theorem is mainly based on Theorem 3 in \cite{JoulaniGS20} and Lemma \ref{lm:surrogate_l} in this paper. We assume
\begin{assumption}
\label{ass:beta_le_beta}
$\|\beta^{t-1}_j \hat{\bm{x}}^{t}_j\|^2_2 \leq  \|\beta^t_j\hat{\bm{x}}^{t+1}_j\|^2_2$ for all $j\in \mathcal{J}$.
\end{assumption}
This assumption can be naturally satisfied after reparameterizing $\beta^t_j$, will be discussed later. If the assumption is true, we have $R^T \leq \sum_{j\in \mathcal{J}}{\left(\beta^{T}_j + \sum_{t=1}^T{ {\|\hat{\bm{r}}^t_j\|^2_2}/{(2\beta^{t}_j}) }\right)}$. In this case, there are many ways to bound the total regret of FD-FTRL (FD-OMD) by $O(\sqrt{T})$. For example, we can set $\beta^t_j= \Theta(\sqrt{t})$, or $\beta^t_j= \Theta(\sqrt{T})$ if $T$ is known. 
Otherwise, we can set $\beta^t_j =\sqrt{\sum_{k=1}^t{\|\hat{\bm{r}}^k_j\|^2_2}}$ to obtain $R^T \leq 2\sum_{j\in \mathcal{J}}\sqrt{\sum_{k=1}^t{\|\hat{\bm{r}}^k_j\|^2_2}}$ \cite{McMahanS10}.

\subsection{Equivalence Theorem}

In this subsection, we prove the equivalence between FD-FTRL (FD-OMD) and CFR-RM (CFR-RM+).
According to Example \ref{ex:2}, it is natural to conjecture that $\alpha^t_j= \sum_{k=1}^t \langle \hat{\bm{l}}^k_{j}, \hat{\bm{x}}^k_{j}\rangle$ \textit{at every decision point} when $\beta^t_j = \|\hat{\bm{R}}^t_j\|_1$. In other words, we may have $\hat{\bm{x}}^{t+1}_j = {[\sum_{k=1}^t \langle \hat{\bm{l}}^k_{j}, \hat{\bm{x}}^k_{j}\rangle\bm{1} - \hat{\bm{L}}^t_j]^+}/{\|\hat{\bm{R}}^t_j\|_1}$ and $\hat{\bm{L}}'^t_j = \hat{\bm{L}}^t_j$, which are exactly the local updates of CFR-RM. It turns out that this conjecture can be proven recursively. 

\begin{theorem}
\label{th:CFR_eq_FTRL}
CFR-RM (CFR-RM+) is equivalent to a special case of FD-FTRL (FD-OMD) with $\beta^t_j = \|[\hat{\bm{R}}^t_j]^+\|_1$ ($\|\hat{\bm{Q}}^t_j\|_1$), $\forall j \in \mathcal{J}, t \geq 0$.
\end{theorem}

The proof is given in Appendix \ref{sec:Equivalence}. We name the FD-FTRL with $\beta^t_j = \|[\hat{\bm{R}}^t_j]^+\|_1$ as FD-FTRL(CFR) and the FD-OMD  with $\beta^t_j = \|\hat{\bm{Q}}^t_j\|_1$ as FD-OMD(CFR). From different perspectives, the equivalences indicate that:
\begin{itemize}
    \item  CFR-RM (CFR-RM+) is an excellent adaptive FTRL (OMD). As mentioned before, we can guarantee $R^T \leq 2\sum_{j}\sqrt{\sum_{k=1}^t{\|\hat{\bm{r}}^k_j\|^2_2}}$ in a case. However, when $\beta^t_j = \|[\hat{\bm{R}}^t_j]^+\|_1$ ($\|\hat{\bm{Q}}^t_j\|_1$), CFR-RM (CFR-RM+) is recovered, and we have $R^T \leq \sum_{j}\max_a[\hat{\bm{R}}^T_j]^+$,\footnote{Can be deduced from Lemma \ref{lm:surrogate_l}, but not from Theorem \ref{th:fd_ada_FTRL}.} which is an even better regret bound as $\max_a[\hat{\bm{R}}^T_j]^+ \leq \sqrt{\sum_{k=1}^t{\|\hat{\bm{r}}^k_j\|^2_2}}$ \cite{CFR}. Since adaptive FTRL (OMD) can adapt to the losses, they are generally faster than the non-adaptive versions. This may partially explain the superior performance of CFRs.
    \item FTRL and OMD can perform as well as CFRs in practice if the parameters are configured properly. While setting $\beta^t_j$ to $\|[\hat{\bm{R}}^t_j]^+\|_1$ or $\|\hat{\bm{Q}}^t_j\|_1$ is unpractical (as this requires running a CFR-RM or a CFR-RM+ in parallel), let $\beta^t_j \approx \|[\hat{\bm{R}}^t_j]^+\|_1$ or $\|\hat{\bm{Q}}^t_j\|_1$ may still result in fast algorithms. Note that this intuition is not restricted to FD-FTRL and FD-OMD, and may apply to general FTRL and OMD with Euclidean regularizers.
\end{itemize}

\subsection{Practical Implementations of FD-FTRL, FD-OMD}
\label{sec:new_CFR}
As we have mentioned, the total regret of FD-FTRL is bounded by $O(\sqrt{T})$ after $T$ iterations if 1) Assumption \ref{ass:beta_le_beta} is fulfilled; and 2) $\beta^t_j = \Theta(\sqrt{t})$ (or $\Theta(\sqrt{T})$). However, Assumption \ref{ass:beta_le_beta} is non-trivial, as it depends on the future decision $\hat{\bm{x}}^{t+1}_j$ (we need to set $\beta^t_j$ before we compute $\hat{\bm{x}}^{t+1}_j$). Alternatively, we introduce a new parameter $\lambda^t_j > 0$ and reparameterize $\beta^t_j$ as ${\sqrt{\lambda^t_j}}/{\|\hat{\bm{x}}^{t+1}_j\|_2}$. Consequently, Assumption \ref{ass:beta_le_beta} can be reformulated as
\begin{assumption}
\label{ass:lambda_le_lambda}
$\lambda^{t-1}_j \leq  \lambda^t_j$ for all $j\in \mathcal{J}$.
\end{assumption}

\textbf{FD-FTRL}.
Recall that in FD-FTRL,
the next decision is computed according to $\hat{\bm{x}}^{t+1}_j =  {[\alpha^t_j\bm{1} - \hat{\bm{L}}'^t_j]^+}/{\beta^t_j}$, where
\begin{equation}
\label{eq:l1_cons}
\alpha^t_j \in  \mathbb{R}, \ s.t.\  \|[\alpha^t_j\bm{1} - \hat{\bm{L}}'^t_j]^+\|_1 = \beta^t_j,
\end{equation}
and $-\psi^{*t}_j(-\hat{\bm{L}}'^t_{j}) = \alpha^t_j$.
Let $\hat{L}'^t_j = -\psi^{*t}_j(-\hat{\bm{L}}'^t_{j})$.
Since $\beta^t_j = {\sqrt{\lambda^t_j}}/{\|\hat{\bm{x}}^{t+1}_j\|_2}$, the constraint becomes 
\begin{equation}
\label{eq:l2_cons}
\|[\hat{L}'^t_j\bm{1} - \hat{\bm{L}}'^t_j]^+\|^2_2 = \lambda^t_j.
\end{equation}

Note that $\hat{L}'^t_j$ in (\ref{eq:l2_cons}) exists and is unique when $\lambda^t_j > 0$. So, it is equal to $\alpha^t_j$ in (\ref{eq:l1_cons}).
Based on the above analysis, we can instantiate an FD-FTRL algorithm that computes the $\hat{L}'^t_j$ at every decision point with respect to (\ref{eq:l2_cons}).
Let $\hat{\bm{R}}'^t_j = \hat{L}'^t_j\bm{1} - \hat{\bm{L}}'^t_j$.
The updates are summarized in Table \ref{tab:comp_algo}, and the algorithm is named \textbf{FD-FTRL(R)}. According to Theorem \ref{th:CFR_eq_FTRL}, when $\lambda^t_j = {\|[\hat{\bm{R}}^t_j]^+\|^2_2}$, which is equivalent to $\beta^t_j = {\|[\hat{\bm{R}}^t_j]^+\|_1}$, FD-FTRL(R) is equivalent to CFR-RM.

\textbf{FD-OMD}.
Recall that FD-OMD has a constraint $\|\hat{\bm{x}}^{t+1}_j\|_1 = 1$, i.e., 
$
\|[\beta^{t-1}_j \hat{\bm{x}}^t_j + \hat{l}'^t_j\bm{1} - \hat{\bm{l}}'^t_j]^+\|_1 = \beta^{t}_j.
$
Since $\beta^t_j = \sqrt{\lambda^t_j}/\|\hat{\bm{x}}^{t+1}_j\|_2$, the constraint is equivalent to 
\begin{equation}
\label{eq:l2_cons_FD-OMD(R)}
\Big\|\Big[{\sqrt{\lambda^{t-1}_j}}\hat{\bm{x}}^t_j/{\|\hat{\bm{x}}^{t}_j\|_2} + \hat{l}'^t_j\bm{1} - \hat{\bm{l}}'^t_j\Big]^+\Big\|^2_2 = \lambda^t_j.
\end{equation}
Note that $\hat{l}'^t_j$ in (\ref{eq:l2_cons_FD-OMD(R)}) exists and is unique. 
Let $\hat{\bm{Q}}'^t_j = \beta^t_j \hat{\bm{x}}^{t+1}_j = [\beta^{t-1}_j \hat{\bm{x}}^t_j + \hat{l}'^t_j\bm{1} - \hat{\bm{l}}'^t_j]^+$.
The updates with respect to (\ref{eq:l2_cons_FD-OMD(R)}) is summarized in Table \ref{tab:comp_algo}, and the algorithm is named \textbf{FD-OMD(R)}. When $\lambda^t_j = \|\hat{\bm{Q}}^t_j\|^2_2$, which is equivalent to $\beta^t_j = \|\hat{\bm{Q}}^t_j\|_1$, according to Theorem \ref{th:CFR_eq_FTRL}, we have FD-OMD(R) equivalent to CFR-RM+.

Notably, FD-FTRL(R) recovers an algorithm named ReCFR \cite{Recursive_CFR}, which is inspired by a warm starting CFR algorithm \cite{StrategyBase}. So, ReCFR is also a special case of FD-FTRL.

As we mentioned before, we may let $\beta^t_j \approx \|[\hat{\bm{R}}^t_j]^+\|_1$ or $\|\hat{\bm{Q}}^t_j\|_1$ to construct a fast FD-FTRL or FD-OMD. Notice that $\|[\hat{\bm{R}}^t_j]^+\|^2_2$ $(\|\hat{\bm{Q}}^t_j\|^2_2) \leq  \sum_{k=1}^t\|\hat{\bm{r}}^k_j\|^2_2$ \cite{CFR,CFRPlus}, we proposed to set $\lambda^t_j$ such that $\lambda^t_j \geq \eta \sum_{k=1}^t{\|\hat{\bm{r}}^k_j\|^2_2}$, $\eta \in (0, \infty)$.
\begin{corollary}
\label{co:FD-FTRL(R)_bound}
If $\eta\sum_{k=1}^t{\|\hat{\bm{r}}^k_j\|^2_2} \leq \lambda^t_j$ and $\lambda^{t-1}_j \leq \lambda^t_j$, $\forall j\in \mathcal{J}$, $t > 0$, then, the total regret of $T$ iterations of FD-FTRL(R) (FD-OMD(R)) is bounded by 
$
\sum_{j\in \mathcal{J}}{\left(\sqrt{n_j} + \frac{1}{\eta}\right)\sqrt{\lambda^T_j}}
$.
\end{corollary}

The proof is given in Appendix \ref{sec:Equivalence}. Clearly, the total regret of FD-FTRL(R) (FD-OMD(R)) is $O(\sqrt{T})$ if $\lambda^t_j = O(t)$ (or $O(T)$), which is achievable as $\|\hat{\bm{r}}^k_j\|^2_2 = O(L^2)$.
Both FD-FTRL(R) and FD-OMD(R) have the same space complexity as vanilla CFR. However, FD-FTRL(R) and FD-OMD(R) need to solve a piecewise constraint at every decision point, which has a time complexity of $O(n^2_j)$. If binary search and Quick Sort are used, an $O(n_j \log n_j)$ complexity can be obtained. Therefore, FD-FTRL(R) and FD-OMD(R) are $|\mathcal{A}|$ or $\log|\mathcal{A}|$ times more expansive than CFRs.

To some extent, FD-FTRL(R) and FD-OMD(R) can be considered as the extensions of CFR-RM and CFR-RM+, respectively. However, they are able to leverage theoretical results in OCO, since they belong to FTRL and OMD families. Another advantage of them over CFRs is that they do not track the counterfactual regrets. 
Note that, FD-FTRL(R) does not track the cumulative loss in practice, too. This is because $\bm{L}^t = \sum_k^t{A\bm{y}^k} =  tA \bar{\bm{y}}^t$ only depends on the average strategy of the opponent, which is available during the process. The disadvantage of FD-FTRL(R) and FD-OMD(R) is that they have a set of parameters. In the next section, we will try two configurations for these parameters.

\begin{figure*}[t]
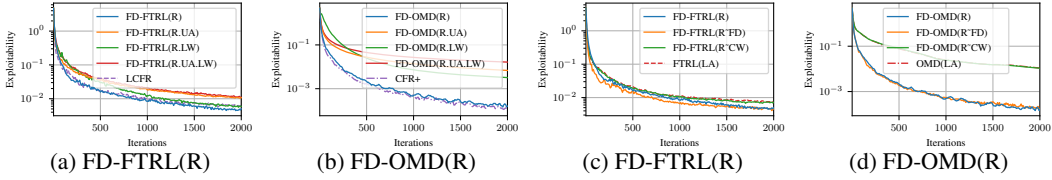

	\centering
    \begin{subfigure}[b]{0.24\linewidth}
        \centering
        \resizebox{\linewidth}{!}{\input{images/plotleduc_poker_ave_compare_iter_13.pgf}}
        \vspace*{-0.25in}
        \label{fig:leduc_FD-FTRL(R)_comp2}
        \caption{FD-FTRL(R)}
    \end{subfigure}
    \hfill
    \begin{subfigure}[b]{0.24\linewidth}
        \centering
        \resizebox{\linewidth}{!}{\input{images/plotleduc_poker_ave_compare_iter_14.pgf}}
        \vspace*{-0.25in}
        \label{fig:leduc_FD-OMD(R)_comp2}
        \caption{FD-OMD(R)}
    \end{subfigure}
    \hfill
    \begin{subfigure}[b]{0.24\linewidth}
        \centering
        \resizebox{\linewidth}{!}{\input{images/plotleduc_poker_ave_compare_iter_11.pgf}}
        \vspace*{-0.25in}
        \label{fig:leduc_FD-FTRL(R)_comp1}
        \caption{FD-FTRL(R)}
    \end{subfigure}
    \hfill
    \begin{subfigure}[b]{0.24\linewidth}
        \centering
        \resizebox{\linewidth}{!}{\input{images/plotleduc_poker_ave_compare_iter_12.pgf}}
        \vspace*{-0.25in}
        \label{fig:leduc_FD-OMD(R)_comp1}
        \caption{FD-OMD(R)}
    \end{subfigure}
    \caption{Exploitability curves of FD-FTRL(R) and FD-OMD(R) in different configurations in Leduc. The x-axis is the number of iterations. \textbf{(a, b)}: weighting methods and averaging methods.
    \textbf{(c, d)}: FD regularizer and \textbf{CW}. FD-FTRL(R\char126FD) is an FD-FTRL(R) without FD regularizer (i.e., an FTRL(LA) with \textbf{CW}). FD-FTRL(R\char126CW) is an FD-FTRL(R) without \textbf{CW}. FD-OMD(R\char126FD) and FD-OMD(R\char126CW) are configured similarly. }
    \label{fig:leduc_comp_all}
\end{figure*}

\begin{figure*}[t]
	\centering
	\begin{subfigure}[b]{0.24\linewidth}
        \centering
        \resizebox{\linewidth}{!}{\input{images/plotleduc_poker_ave_compare_iter_8.pgf}}
        \vspace*{-0.25in}
        \label{fig:leduc_poker}
        \caption{Leduc}
    \end{subfigure}
    \hfill
    \begin{subfigure}[b]{0.24\linewidth}
        \centering
        \resizebox{\linewidth}{!}{\input{images/plotleduc18_poker_ave_compare_iter_8.pgf}}
        \vspace*{-0.25in}
        \label{fig:leduc18}
        \caption{Leduc(2, 9)}
    \end{subfigure}
    \hfill
    \begin{subfigure}[b]{0.24\linewidth}
        \centering
        \resizebox{\linewidth}{!}{\input{images/plotgoofspiel_4_ave_compare_iter_8.pgf}}
        \vspace*{-0.25in}
        \label{fig:goof_4}
        \caption{Goofspiel 4}
    \end{subfigure}
    \hfill
    \begin{subfigure}[b]{0.24\linewidth}
        \centering
        \resizebox{\linewidth}{!}{\input{images/plotgoofspiel_4_imp_ave_compare_iter_8.pgf}}
        \vspace*{-0.25in}
        \label{fig:goop_4_imp}
        \caption{Goofspiel 4(imp.)}
    \end{subfigure}
    \\
    \begin{subfigure}[b]{0.24\linewidth}
        \centering
        \resizebox{\linewidth}{!}{\input{images/plotFHP2_poker_ave_compare_iter_8.pgf}}
        \vspace*{-0.25in}
        \label{fig:FHP_2_5}
        \caption{FHP(2, 5)}
    \end{subfigure}
    \hfill
    \begin{subfigure}[b]{0.24\linewidth}
        \centering
        \resizebox{\linewidth}{!}{\input{images/plotgoofspiel_5_ave_compare_iter_8.pgf}}
        \vspace*{-0.25in}
        \label{fig:Goofspiel}
        \caption{Goofspiel}
    \end{subfigure}
    \hfill
    \begin{subfigure}[b]{0.24\linewidth}
        \centering
        \resizebox{\linewidth}{!}{\input{images/plotliars_dice_ave_compare_iter_8.pgf}}
        \vspace*{-0.25in}
        \label{fig:liars_dice}
        \caption{Liar's dice}
    \end{subfigure}
    \hfill
    \begin{subfigure}[b]{0.24\linewidth}
        \centering
        \resizebox{\linewidth}{!}{\input{images/plotbattleship_3_2_3_ave_compare_iter_8.pgf}}
        \vspace*{-0.25in}
        \label{fig:battleship}
        \caption{Battleship}
    \end{subfigure}
    \\
    \begin{subfigure}[b]{\linewidth}
        \centering
        \includegraphics[height=0.25in]{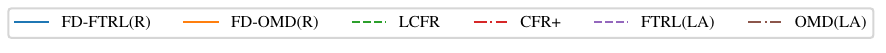}
        \vspace*{-0.1in}
    \end{subfigure}
    \caption{Exploitability curves of FD-FTRL(R), FD-OMD(R), and the competitors in eight games.}
    \label{fig:comp_all}
\end{figure*}

\section{Experimental Investigation}

To better understand the properties of FD-FTRL(R) and FD-OMD(R), we test two different methods for setting the weighting parameters $\lambda^t_j$: 1) \textbf{Linear Weighting (LW)}: $\lambda^t_j = \eta \overline{y}^t_j n_j L^2 t$ \cite{Recursive_CFR}; 2) \textbf{Constant Weighting (CW)}\footnote{Since $\overline{y}^t_j$ changes slowly, we can consider it constant.}: $\lambda^t_j = {\eta \overline{y}^t_j} n_j L^2 T $.
In the equations, $y^t_j$ denotes the probability of reaching $j$ of the opponent, and $\overline{y}^t_j$ is the average probability. $\eta \in (0, \infty)$ is a global hyper-parameter. 
These configurations are inspired by the observations that $\|\hat{\bm{r}}^t_j\|^2_2 \leq 4y^t_j n_j L^2$ and $ \sum_{k=1}^t\|\hat{\bm{r}}^k_j\|^2_2 \leq 4\overline{y}^t_j n_j L^2 t$ \cite{StrategyBase}. According to Corollary \ref{co:FD-FTRL(R)_bound}, FD-FTRL(R) and FD-OMD(R) with both weighting methods have a sub-linear regret bound $O(\sqrt{T})$.
For computing the average strategy, we use: 1) \textbf{Uniform Averaging (UA)}: $\overline{\bm{x}}^{T} = \frac{1}{T}\sum_{t=1}^{T}{\bm{x}^t}$; 2) \textbf{Linear Averaging (LA)}: $\overline{\bm{x}}^{T} = \frac{2}{T(T+1)}\sum_{t=1}^{T}{t\bm{x}^t}$. FD-FTRL(R) and FD-OMD(R) use \textbf{CW} and \textbf{LA} by default. 

The algorithms are compared with FTRL, OMD and CFRs.
For a fair comparison, the competitors also use \textbf{LA} for computing the average strategy, namely Linear CFR (LCFR) \cite{discfr}, CFR+, FTRL(LA), and OMD(LA). FTRL(LA) and OMD(LA)\footnote{A primary experiment shows that the \textbf{LA} versions of FTRL and OMD are significantly faster than the \textbf{UA} versions.} are the algorithms with \textbf{LA} and a regularizer $q^{0:t}(\bm{x}) = q^{0}(\bm{x}) = \sum_{j\in\mathcal{J}}\bm{x}[p_j] \frac{1}{2} \beta_j \|\hat{\bm{x}}\|^2_2$ for $t > 0$. We set $\{\beta_j\}$ in FTRL(LA) and OMD(LA) according to \cite{OptimisticRegret}: $\beta_j = 2\sigma + 2 \max_{a\in A_j}\sum_{j'\in C_{j,a}}\beta_{j'}$, where $\sigma$ is a hyper-parameter.
In FD-FTRL(R) and FTRL(LA), the losses are also weighted linearly, as in LCFR \cite{discfr}. All the algorithms use alternating updates, as in CFRs.

We run a coarse hyper-parameter tuning for each algorithm, and the best results are reported. The tuning results of FD-FTRL(R) and FD-OMD(R) are given in Appendix \ref{sec:results}.
We conduct our experiments in eight benchmark games, including Leduc \cite{Leduc} and FHP \cite{DeepCFR}. A description of the games is given in Appendix \ref{sec:descript}. The experiments use part of the code of project OpenSpiel \cite{openspiel} .\footnote{https://github.com/deepmind/open\_spiel}
The license is Apache-2.0.

We first compare FD-FTRL(CFR) and FD-OMD(CFR) with vanilla CFR and CFR+ in the benchmark games. The results are given in Appendix \ref{sec:results}, showing that the exploitability curves of FD-FTRL(CFR) (FD-OMD(CFR)) and vanilla CFR (CFR+) are overlapping with each other in every game. Therefore, the equivalences are also verified experimentally. Then, we test FD-FTRL(R) and FD-OMD(R) in different configurations. The results in Leduc are given in Figure \ref{fig:leduc_comp_all}. 
In plot \ref{fig:leduc_comp_all}(a), the results show that FD-FTRL(R.LW) and FD-FTRL(R.UA) are slower than the default FD-FTRL(R), which indicates that both \textbf{CW} and \textbf{LA} contribute to the performance of FD-FTRL(R). 
In plot \ref{fig:leduc_comp_all}(b), as we can see, the default FD-OMD(R) with \textbf{LA} and \textbf{CW} is much faster. 
This is not surprising, as it is also observed in CFR+ that \textbf{LA} can improve the performance significantly \cite{CFRPlus}. However, the reason is still not clear. 
In plots \ref{fig:leduc_comp_all}(c, d), we perform an ablation study for FD-FTRL(R) and FD-OMD(R). The results show that \textbf{CW} has a significant impact on the performance, while FD regularizer has a much weaker effect. Since FD-FTRL(R\char126FD) is essentially an FTRL(LA) with \textbf{CW}, the results also indicate that \textbf{CW} can apply to conventional FTRL algorithms. The results in the other games are reported in Appendix \ref{sec:results}, which are basically consistent with the results in Leduc. It is worth noting that FD-FTRL(R\char126FD) is faster than FD-FTRL(R) in some benchmark games. 
This suggests that FD-FTRL(R) (as well as vanilla CFR and LCFR) may be too conservative in some games.
However, FD-OMD(R) is always (one of) the fastest. 

The results of FD-FTRL(R) and FD-OMD(R) in all the benchmark games are shown in Figure \ref{fig:comp_all}. As we can see, FD-FTRL(R) is tied with LCFR and FTRL(LA). However,
the results show that FD-OMD(R) behaves more like a CFR instead of an OMD: it is always faster than LCFR and OMD(LA), and only slower than CFR+ in Leduc(2, 9) and FHP(2, 5).
Note that Leduc(2, 9) and FHP(2, 5) are the most stochastic among the benchmark games. So, it seems that the weighting methods (\textbf{CW} and \textbf{LW}) can not handle stochastic games well.
We suspect FD-FTRL(R) and FD-OMD(R) may perform better in stochastic games when $\lambda^t_j$ is set closer to $\|[\hat{\bm{R}}^t_j]^+\|^2_2$ and $\|\hat{\bm{Q}}^t_j\|^2_2$.

We also compare FD-FTRL(R) and FD-OMD(R) with PCFR and PCFR+ \cite{Blackwell}. The results are given in Appendix \ref{sec:results}, showing that FD-FTRL(R) and FD-OMD(R) are also competitive compared to them. Finally, we compare FD-FTRL(R) and FD-OMD(R) with FTRL, OMD, and vanilla CFR that use \textbf{UA}. The results are given in Appendix \ref{sec:results}, showing that both FD-FTRL(R) and FD-OMD(R) are always faster than them.

\section{Conclusions and Future Work}
In the paper, It is proven that CFR-RM and CFR-RM+ are equivalent to special cases of FTRL and OMD, respectively. As the bridges, FD-FTRL and FD-OMD have been proposed, and have been extensively tested in eight benchmark EFGs. The experimental results show that they are competitive compared to conventional FTRL, OMD, and CFRs. The results also suggest that FTRL and OMD are not necessarily slower than CFRs in EFGs. Therefore, more research is required in applying OCO algorithms to EFGs.

The equivalence analysis in this paper is relatively primitive, e.g., the analysis is restricted to RM, RM+, and Euclidean regularizers. Also, the analysis does not improve the theoretical results of CFR and CFR+. However, it can explain CFRs as an adaptive FTRL (OMD) and provides intuitions in applying FTRL and OMD to EFGs. Besides, since there are optimistic variants of FTRL and OMD \cite{OptimisticRegret} that converge at a rate of $O(1/T)$, we may also be able to develop new optimistic variants of CFRs.

In recent years, function-approximate CFRs \cite{RCFR,DeepCFR,DoubleCFR,Recursive_CFR} have been found to have problems in approximating cumulative counterfactual regrets. Combining FD-FTRL(R) or FD-OMD(R) with function approximation would not have these problems, since they do not rely on the cumulative counterfactual regrets. Furthermore, It is known that CFR-RM+ works much better with function approximation than CFR-RM \cite{morrill2016using,Orazio2020}. It would be interesting to see whether FD-OMD(R) is more favorable than FD-FTRL(R) in this setting.

\clearpage

\bibliography{cfr_eq_oco}

\newpage
\appendix
\onecolumn

\section{Notation Table}
\label{sec:notation}
In this section, we list the notations that appear in the article. See table \ref{tab:Notation}. We mainly follow the notations in \cite{OptimisticRegret} and \cite{Blackwell}. However, as this paper discusses the equivalences between algorithms in different areas, it is unavoidable to introduce new notations. Also, some notations may not be consistent with the ones defined in the mentioned literature. For example, we use $\bm{x}_j$ to denote a sequence-form strategy in $\mathcal{X}_j$, while \cite{OptimisticRegret} uses it to denote a sub-vector of $\bm{x}$ that corresponds to decision point $j$.
\begin{table}[htbp]
		\centering
		\caption{Notation Table}
		\label{tab:Notation}
		\begin{tabular}{cl}
		\toprule
			$\mathcal{J}$ & the set of decision points. \\
			${j}$& $j\in \mathcal{J}$. A decision point. \\
			$\mathcal{K}$ & the set of observation points. \\
			${k}$ & $k\in \mathcal{K}$. An observation point. \\
			${A}_j$ & the action set at decision point $j$. \\ 
			$a$ & $a\in A_j$. An action in $A_j$. \\
			$\mathcal{A}$ & the set of all actions. \\
			$\bm{n}_j$ & $\bm{n}_j = |{A}_j|$, the size of the action set at decision point $j$. \\
			$C_{j, a}$ & the set of earliest reachable decision points after taking action $a$ at $j$. \\
			$C_{\downarrow j}$& the set of all decision points reachable from $j$, including $j$. \\
			$o$ & the root decision point. \\
			\midrule
			$\bm{1}$ & an all-ones vector. \\
			$\D^n$ & an $n$-dimensional simplex. \\
			$\hat{\bm{x}}_j$ & $\hat{\bm{x}}_j \in \D^{n_j}$. A decision at point $j$. \\
			$\mathcal{X}$ & the sequence-form strategy space (of player 1). \\
			$\mathcal{X}_j$ & the sequence-form strategy space in a sub-sequential decision process starting from $j$. \\
			$\mathcal{Y}$ & the sequence-form strategy space of player 2. \\
			$\bm{x}$ & $\bm{x} \in \mathcal{X}$. A sequence-form strategy of player 1. \\
			$\bm{x}_j$ & $\bm{x}_j \in \mathcal{X}_j$. A sequence-form strategy of player 1 in a sub-game. \\
			$\overline{\bm{x}}$& the average strategy of player 1. \\
			$\bm{y}$& $\bm{y} \in \mathcal{Y}$. A sequence-form strategy of player 2. \\
			$\overline{\bm{y}}$& the average strategy of player 2. \\
			$\bm{x}[p_{j}]$ & the entry in $\bm{x}$ corresponding to the sequence of reaching $j$. \\
			$\bm{z}[j]$ & the $n_j$ entries related to $j$ for any $\bm{z}\in \mathbb{R}^{\sum_{k=1}^m{n_{j_k}}}$. \\
			$\bm{z}[j, a]$ & the entry corresponding to $(j, a)$ for any vector $\bm{z}\in \mathbb{R}^{\sum_{k=1}^m{n_{j_k}}}$. \\
			$\bm{z}[a]$ & the entry corresponding to action $a$ for any vector $\bm{z}\in \mathbb{R}^{n_j}$. \\
			\midrule
			$\bm{l}$ &  the loss vector. $\bm{l} = \mathbf{A}\bm{y}$ in a two-player zero-sum game. \\
			$\hat{\bm{l}}^t_j$ &  $\hat{\bm{l}}^t_j[a] = \bm{l}^t[j, a] + \sum_{j'\in C_{j, a}}{\langle \hat{\bm{l}}^t_{j'}, \hat{\bm{x}}^t_{j'}\rangle}$. The counterfactual loss. \\
			$\bm{L}^T$ & $\bm{L}^T = \sum_{t=1}^T \bm{l}^t$. The cumulative loss. \\
			$\hat{\bm{L}}^T_j$ & $\hat{\bm{L}}^t_j = \sum_{t=1}^T \hat{\bm{l}}^t_j$. The cumulative counterfactual loss at $j$. \\
			$\hat{\bm{r}}^t_j$ & $\hat{\bm{r}}^t_j =  \langle \hat{\bm{l}}^t_{j}, \hat{\bm{x}}^t_{j} \rangle\bm{1} - \hat{\bm{l}}^t_j$. The instantaneous counterfactual regret. \\
			$\hat{\bm{R}}^t_j$ & $\hat{\bm{R}}^t_j = \sum_{k=1}^{t}\hat{\bm{r}}^k_j = \sum_{k=1}^t \langle \hat{\bm{l}}^k_{j}, \hat{\bm{x}}^k_{j}\rangle\bm{1} - \hat{\bm{L}}^t_j$. The cumulative counterfactual regret. \\
			$\hat{R}^T_j$ & $\hat{R}^T_j = \max_{\hat{\bm{x}}'_j \in \D^{n_j}}\sum_{t=1}^{T}{\langle \hat{\bm{r}}^t_j, \hat{\bm{x}}'_j\rangle} = \max_{a\in A_j}\hat{\bm{R}}^T_j[a]$. The immediate counterfactual regret. \\
			$\hat{\bm{Q}}^{t}_j$ & $\hat{\bm{Q}}^{t}_j \gets [\hat{\bm{Q}}^{t-1}_j + \langle \hat{\bm{l}}^t_{j}, \hat{\bm{x}}^t_{j} \rangle\bm{1} -  \hat{\bm{l}}^t_j]^+$. The cumulative counterfactual regret of CFR-RM+. \\
			\midrule
			$q^{0:t}(\bm{x})$ & $q^{0:t}(\bm{x}) = \sum_{k=1}^t{q^k(\bm{x})}$. The regularizer. $q^k: D \to \mathbb{R}, \mathcal{X} \subseteq D$ is a proper function. \\
			$d(\bm{x})$ & $d(\bm{x})= \sum_{j\in \mathcal{J}}{\bm{x}[p_j]\psi_j\left(\hat{\bm{x}}_j\right)}$. A dilated DGF. $\psi_j: E \to \mathbb{R}, \D^{n_j} \subseteq E$. \\
			$\mathcal{B}_{d}(\bm{x}'||\bm{x})$ & $\mathcal{B}_{d}(\bm{x}'||\bm{x}) = d(\bm{x}') -  d(\bm{x}_{j}) - \left\langle \nabla  d(\bm{x}), \bm{x}' - \bm{x} \right\rangle$. The Bregman divergence of a DGF $d$. \\
			\midrule
			$\hat{\bm{L}}'^t_j$ & $\hat{\bm{L}}'^t_j[a] = \bm{L}^t[j, a] + \sum_{j'\in C_{j, a}}{-\psi^{*t}_{j'}(-\hat{\bm{L}}'^t_{j'})}$. The local loss of FTRL. \\
			$\hat{L}'^t_j$ & $\hat{L}'^t_j = -\psi^{*t}_{j}(-\hat{\bm{L}}'^t_{j})$. \\
			$\hat{\bm{l}}'^t_j$ & $\hat{\bm{l}}'^t_j[a] =  \bm{l}^t[j, a] + \sum_{j'\in C_{j, a}}{\hat{l}'^t_{j'}}$. The local loss of OMD. \\
			$\hat{l}'^t_{j}$ & $\hat{l}'^t_{j} = {\psi}^{*t-1}_{j}(\nabla \psi^{t-1}_{j}(\hat{\bm{x}}^{t}_{j}))-{\psi}^{*t}_{j}(\nabla \psi^{t-1}_{j}(\hat{\bm{x}}^{t}_{j}) -\hat{\bm{l}}'^t_{j})$. \\
			$\hat{\bm{R}}'^t_j$ & $\hat{\bm{R}}'^t_j = \hat{L}'^t_j\bm{1} - \hat{\bm{L}}'^t_j$. \\
             $\hat{\bm{Q}}'^t_j$ & $\hat{\bm{Q}}'^t_j = \beta^t_j \hat{\bm{x}}^{t+1}_j = [\beta^{t-1}_j \hat{\bm{x}}^t_j + \hat{l}'^t_j\bm{1} - \hat{\bm{l}}'^t_j]^+$. \\
			\midrule 
			$\beta^t_j$ & $\beta^t_j > 0$. The weighting parameter for DGF $\psi^{t}_j$. \\
			$\lambda^{t}_j$ & $\lambda^t_j > 0$. The weighting parameter for reparameterizing $\beta^t_j$: $\beta^t_j= {\sqrt{\lambda^t_j}}/{\|\hat{\bm{x}}^{t+1}_j\|_2}$. \\
		\bottomrule
		\end{tabular}
	\end{table}
	
\section{An illustration of Sequential Decision Process}
\label{sec:illustration}
\begin{figure}[htp]
    \centering
    \includegraphics[width=0.5\linewidth]{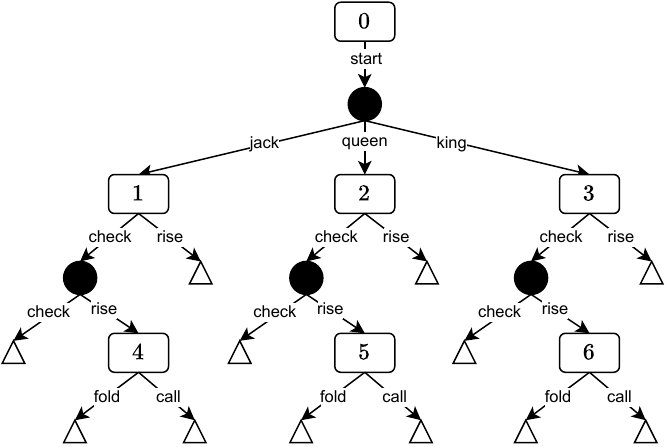}
    \caption{Sequential action space for the first player in the game of Kuhn poker, $\bullet$ denotes an observation point, $\triangle$ denotes the end of decision process. Adapted from \cite{OptimisticRegret}.}
    \label{fig:kuhn}
\end{figure}

As an illustration, consider the game of Kuhn
poker. Kuhn poker consists of a three-card deck:
king, queen, and jack. The action space for the
first player is shown in Figure \ref{fig:kuhn}. For example, we have decision set $\mathcal{J}=\left\{0,1,2,3,4,5,6  \right\}$, and action sets ${A}_0=\left\{start \right\}$, ${A}_1= {A}_2= {A}_3= \left\{check, raise \right\}$, and ${A}_4={A}_5={A}_6=\left\{fold, call\right\}$. Moreover, we have  ${C}_{0,start}=\left\{1,2,3\right\}$, ${C}_{1,raise}=\emptyset$, ${C}_{3,check}=\left\{6\right\}$, $C_{\downarrow 0} = \mathcal{J}$, $C_{\downarrow 1} = \left\{1, 4\right\}$, and et al.

For a sequence-form strategy $\bm{x}\in \mathcal{X}$ in Kuhn poker,
we have an $\bm{x}_{6}$ in $\mathcal{X}_{6}$ equals $\hat{\bm{x}}_6$, an ${\bm{x}}_{3}$ in $\mathcal{X}_{3}$ equals $(\hat{\bm{x}}_3, \hat{\bm{x}}_{3}[check] {\bm{x}}_{6}) = (\hat{\bm{x}}_3, \hat{\bm{x}}_{3}[check] \hat{\bm{x}}_6)$, and
\begin{equation*}
\begin{aligned}
    \bm{x} = & \left(\hat{\bm{x}}_{o},  \hat{\bm{x}}_{o}[start]\bm{x}_{1}, \hat{\bm{x}}_{o}[start]\bm{x}_{2}, \hat{\bm{x}}_{o}[start]\bm{x}_{3}\right) \\
    = & \left(1,  \hat{\bm{x}}_1, \hat{\bm{x}}_{1}[check] \hat{\bm{x}}_4, \hat{\bm{x}}_2, \hat{\bm{x}}_{2}[check] \hat{\bm{x}}_5, \hat{\bm{x}}_3, \hat{\bm{x}}_{3}[check] \hat{\bm{x}}_6\right).
\end{aligned}
\end{equation*}

\section{Counterfactual Regret Minimization}
\label{sec:CFR}

\subsection{A discussion on Assumption \ref{ass:R_le_R}}
CFR usually initializes $\hat{\bm{R}}^{0}_j$ or $\hat{\bm{Q}}^{0}_j$ to zeros. 
In this paper, we set  $\hat{\bm{R}}^{0}_j$ or $\hat{\bm{Q}}^{0}_j$ to a small value vector $\epsilon \bm{1} > 0$ for each $j\in \mathcal{J}$\footnote{This setting has been adopted in OpenSpiel project \cite{openspiel}: https://github.com/deepmind/open\_spiel}. As shown in Lemma \ref{lm:CFR_FTRL_R_bound} and \ref{lm:CFR_OMD_Q_bound}, the $l_2$-norm of the cumulative counterfactual regret at every decision point is monotonically increasing. So, \textit{it is guaranteed that $\|[\hat{\bm{R}}^{t}_j]^+\|_1 > 0$ and $\|\hat{\bm{Q}}^{t}_j\|_1 > 0$.} Note that, according to Lemma \ref{lm:CFR_FTRL_R_bound}, this initialization has a negligible effect on the convergence, as $\hat{R}^T_j \leq \|[\hat{\bm{R}}^T_j]^+\|_2 \leq \sqrt{\epsilon^2n_j + \sum_{t=1}^T{ \|\hat{\bm{r}}^t_j\|^2_2}} =  O(\sqrt{T})$. The effect on CFR-RM+ is similar.

\begin{lemma}
\label{lm:CFR_FTRL_R_bound}
In CFR-RM, 
$\|[\hat{\bm{R}}^{t-1}_j]^+\|^2_2 \leq \|[\hat{\bm{R}}^{t}_j]^+\|^2_2 \leq \|[\hat{\bm{R}}^{t-1}_j]^+\|^2_2 + \|\hat{\bm{r}}^{t}_j\|^2_2$  for all $j \in \mathcal{J}$ and $t > 0$.
\end{lemma}

\begin{proof}
First, we prove that $\|[\hat{\bm{R}}^{t-1}_j]^+\|^2_2 \leq \|[\hat{\bm{R}}^{t}_j]^+\|^2_2$.
It is trivial when $\|[\hat{\bm{R}}^{t-1}_j]^+\|^2_2 = 0$.
When $\|[\hat{\bm{R}}^{t-1}_j]^+\|^2_2 > 0$, we have,
\begin{equation}
\label{eq:raw_R_eq}
\begin{aligned}
\langle [\hat{\bm{R}}^{t}_{j}]^+, [\hat{\bm{R}}^{t-1}_{j}]^+ \rangle \geq &  \langle \hat{\bm{R}}^{t}_{j}, [\hat{\bm{R}}^{t-1}_{j}]^+ \rangle \\
= & \langle \hat{\bm{R}}^{t - 1}_{j} + \hat{\bm{r}}^t_j, [\hat{\bm{R}}^{t-1}_{j}]^+ \rangle \\
= & \langle \hat{\bm{R}}^{t - 1}_{j}, [\hat{\bm{R}}^{t-1}_{j}]^+ \rangle \\
= & \|[\hat{\bm{R}}^{t-1}_j]^+\|^2_2.
\end{aligned}
\end{equation}
The second equality is because  
\begin{equation}
\begin{aligned}
\langle \hat{\bm{r}}^t_j, [\hat{\bm{R}}^{t-1}_{j}]^+ \rangle = & \langle \hat{\bm{l}}^t_{j}, \hat{\bm{x}}^t_{j} \rangle  \|[\hat{\bm{R}}^{t-1}_{j}]^+\|_1  - \langle \hat{\bm{l}}^t_j, [\hat{\bm{R}}^{t-1}_{j}]^+ \rangle \\
= & \langle \hat{\bm{l}}^t_{j}, \hat{\bm{x}}^t_{j} \rangle  \|[\hat{\bm{R}}^{t-1}_{j}]^+\|_1  - \langle \hat{\bm{l}}^t_j, \hat{\bm{x}}^{t}_{j} \rangle \|[\hat{\bm{R}}^{t-1}_{j}]^+\|_1 \\
= & 0.
\end{aligned}
\end{equation}

Secondly, according to Cauchy–Schwarz inequality, we have
\begin{equation}
\begin{aligned}
\langle [\hat{\bm{R}}^{t}_{j}]^+, [\hat{\bm{R}}^{t-1}_{j}]^+ \rangle \leq & \|[\hat{\bm{R}}^{t-1}_j]^+\|_2 \|[\hat{\bm{R}}^{t}_j]^+\|_2 \\
\leq &  \sqrt{\langle [\hat{\bm{R}}^{t}_{j}]^+, [\hat{\bm{R}}^{t-1}_{j}]^+ \rangle}\|[\hat{\bm{R}}^{t}_j]^+\|_2.
\end{aligned}
\end{equation}
Since $\langle [\hat{\bm{R}}^{t}_{j}]^+, [\hat{\bm{R}}^{t-1}_{j}]^+ \rangle \geq \|[\hat{\bm{R}}^{t-1}_j]^+\|^2_2 > 0$, we have
\begin{equation}
\|[\hat{\bm{R}}^{t-1}_j]^+\|^2_2 \leq  \|[\hat{\bm{R}}^{t}_j]^+\|^2_2.
\end{equation}

Besides, we have 
\begin{equation}
\begin{aligned}
\|[\hat{\bm{R}}^{t}_j]^+\|^2_2 = & 
\|[\hat{\bm{R}}^{t-1}_j + \hat{\bm{r}}^t_j]^+\|^2_2 \\
\leq & 
\|[\hat{\bm{R}}^{t-1}_j]^+ + \hat{\bm{r}}^t_j\|^2_2 \\
= & \|[\hat{\bm{R}}^{t-1}_j]^+\|^2_2 + 2\langle \hat{\bm{r}}^t_j, [\hat{\bm{R}}^{t-1}_{j}]^+ \rangle + \|[\hat{\bm{r}}^t_j]^+\|^2_2 \\
= & \|[\hat{\bm{R}}^{t-1}_j]^+\|^2_2 + \|\hat{\bm{r}}^t_j\|^2_2 .
\end{aligned}
\end{equation}
\end{proof}

\begin{lemma}
\label{lm:CFR_OMD_Q_bound}
In CFR-RM+,
$\|\hat{\bm{Q}}^{t-1}_j\|^2_2 \leq \|\hat{\bm{Q}}^{t}_j\|^2_2 \leq \|\hat{\bm{Q}}^{t-1}_j\|^2_2 + \|\hat{\bm{r}}^{t}_j\|^2_2$ for all $j \in \mathcal{J}$ and $t > 0$.
\end{lemma}

\begin{proof}
First, we prove that $\|\hat{\bm{Q}}^{t-1}_j\|^2_2 \leq  \|\hat{\bm{Q}}^{t}_j\|^2_2$.
It is trivial when $\|\hat{\bm{Q}}^{t-1}_j\|^2_2 = 0$.
When $\|\hat{\bm{Q}}^{t-1}_j\|^2_2 > 0$, we have,
\begin{equation}
\begin{aligned}
\langle \hat{\bm{Q}}^{t}_{j}, \hat{\bm{Q}}^{t-1}_{j} \rangle \geq & \langle \hat{\bm{Q}}^{t - 1}_{j} + \hat{\bm{r}}^t_j, \hat{\bm{Q}}^{t-1}_{j} \rangle \\
= & \langle \hat{\bm{Q}}^{t - 1}_{j}, \hat{\bm{Q}}^{t-1}_{j} \rangle \\
= & \|\hat{\bm{Q}}^{t-1}_j\|^2_2.
\end{aligned}
\end{equation}
The Second equality is because  

\begin{equation}
\begin{aligned}
\langle \hat{\bm{r}}^t_j, \hat{\bm{Q}}^{t-1}_{j} \rangle = &\langle \hat{\bm{l}}^t_{j}, \hat{\bm{x}}^t_{j} \rangle  \|\hat{\bm{Q}}^{t-1}_{j}\|_1  - \langle \hat{\bm{l}}^t_j, \hat{\bm{Q}}^{t-1}_{j} \rangle \\
= & \langle \hat{\bm{l}}^t_{j}, \hat{\bm{x}}^t_{j} \rangle  \|\hat{\bm{Q}}^{t-1}_{j}\|_1  - \langle \hat{\bm{l}}^t_j, \hat{\bm{x}}^{t}_{j} \rangle \|\hat{\bm{Q}}^{t-1}_{j}\|_1 \\
= & 0.
\end{aligned}
\end{equation}

Secondly, according to Cauchy–Schwarz inequality, we have
\begin{equation}
\begin{aligned}
\langle \hat{\bm{Q}}^{t}_{j}, \hat{\bm{Q}}^{t-1}_{j} \rangle \leq & \|\hat{\bm{Q}}^{t-1}_j\|_2 \|\hat{\bm{Q}}^{t}_j\|_2 \\
\leq &  \sqrt{\langle \hat{\bm{Q}}^{t}_{j}, \hat{\bm{Q}}^{t-1}_{j} \rangle}\|\hat{\bm{Q}}^{t}_j\|_2.
\end{aligned}
\end{equation}
So,
\begin{equation}
\|\hat{\bm{Q}}^{t-1}_j\|^2_2 \leq  \|\hat{\bm{Q}}^{t}_j\|^2_2.
\end{equation}

Besides, we have 
\begin{equation}
\begin{aligned}
\|\hat{\bm{Q}}^{t}_j\|^2_2 = & 
\|[\hat{\bm{Q}}^{t-1}_j + \hat{\bm{r}}^t_j]^+\|^2_2 \\
\leq & 
\|\hat{\bm{Q}}^{t-1}_j + \hat{\bm{r}}^t_j\|^2_2 \\
= & \|\hat{\bm{Q}}^{t-1}_j\|^2_2 + 2\langle \hat{\bm{r}}^t_j, \hat{\bm{Q}}^{t-1}_{j} \rangle + \|[\hat{\bm{r}}^t_j]^+\|^2_2 \\
\leq & \|\hat{\bm{Q}}^{t-1}_j\|^2_2 + \|\hat{\bm{r}}^t_j\|^2_2 .
\end{aligned}
\end{equation}

\end{proof}

\subsection{Proof for Lemma \ref{lm:surrogate_l}}

\begin{proof}
As in the paper, the instantaneous counterfactual regret is defined as $\hat{\bm{r}}_{j} = \langle \hat{\bm{l}}_j, \hat{\bm{x}}_{j} \rangle\bm{1} - \hat{\bm{l}}_j$, where
\begin{equation}
\hat{\bm{l}}_j[a] = \bm{l}[j, a] + \sum_{j'\in C_{j, a}}{\langle \hat{\bm{l}}_{j'}, \hat{\bm{x}}_{j'} \rangle}.
\end{equation}

Therefore, 
\begin{equation}
\begin{aligned}
\bm{x}'[p_{j}] \langle  \hat{\bm{l}}_{j}, \hat{\bm{x}}'_{j} \rangle
= &  \bm{x}'[p_{j}]\langle \bm{l}[j], \hat{\bm{x}}'_{j} \rangle + \bm{x}'[p_{j}]\sum_{a\in A_j} \sum_{j'\in C_{j, a}}{ \hat{\bm{x}}'_{j}[a] \langle \hat{\bm{l}}_{j'}, \hat{\bm{x}}_{j'} \rangle }  \\
= &   \bm{x}'[p_{j}]\langle \bm{l}[j], \hat{\bm{x}}'_{j} \rangle  + \sum_{a\in A_j}\sum_{j'\in C_{j, a}}{ \bm{x}'[p_j] \langle \hat{\bm{l}}_{j'}, \hat{\bm{x}}_{j'} \rangle} 
\end{aligned}
\end{equation}
So,

\begin{equation}
\begin{aligned}
 \sum_{j\in\mathcal{J}}{ \bm{x}'[p_{j}] \langle  \hat{\bm{r}}_{j}, \hat{\bm{x}}'_{j} \rangle }  =  & \sum_{j\in\mathcal{J}}{ \bm{x}'[p_{j}] \langle  \hat{\bm{l}}_{j}, \hat{\bm{x}}_{j} \rangle } - \sum_{j\in\mathcal{J}}{ \bm{x}'[p_{j}] \langle  \hat{\bm{l}}_{j}, \hat{\bm{x}}'_{j} \rangle } \\
 = & \sum_{j\in\mathcal{J}}{ \bm{x}'[p_{j}] \langle \hat{\bm{l}}_j, \hat{\bm{x}}_{j} \rangle} - \left\{\sum_{j\in\mathcal{J}}{\bm{x}'[p_{j}]\langle \bm{l}[j], \hat{\bm{x}}'_{j} \rangle} + \sum_{j\in\mathcal{J}}{\sum_{a\in A_j}\sum_{j'\in C_{j, a}}{ \bm{x}'[p_j] \langle \hat{\bm{l}}_{j'}, \hat{\bm{x}}_{j'} \rangle }} \right\} \\
 = & \left\{\sum_{j\in\mathcal{J}}{ \bm{x}'[p_{j}] \langle \hat{\bm{l}}_j, \hat{\bm{x}}_{j} \rangle} - \sum_{j\in\mathcal{J}}{\sum_{a\in A_j}\sum_{j'\in C_{j, a}}{ \bm{x}'[p_j] \langle \hat{\bm{l}}_{j'}, \hat{\bm{x}}_{j'} \rangle }}\right\} - \sum_{j\in\mathcal{J}}{\bm{x}'[p_{j}]\langle \bm{l}[j], \hat{\bm{x}}'_{j} \rangle }  \\
 = & \langle \hat{\bm{l}}_o, \hat{\bm{x}}_o \rangle - \langle \bm{l}, \bm{x}' \rangle \\
 = &  \langle \bm{l}, \bm{x} \rangle - \langle \bm{l}, \bm{x}' \rangle.
\end{aligned}
\end{equation}
The last equality is because $\langle \hat{\bm{l}}_o, \hat{\bm{x}}_{o} \rangle = \langle \bm{l}, \bm{x} \rangle $, according to the definition of counterfactual losses. Rearranging the equation gives the lemma.

\end{proof}

\section{Online Convex Optimization and Distance-Generating Function}
\label{sec:OCO}

\subsection{Recursive Definition of Dilated DGF}

Given a sequence-form space $\mathcal{X}$, a dilated  DGF is defined as 
\begin{equation}
d(\bm{x})= \sum_{j\in \mathcal{J}}{ \bm{x}[p_j]\psi_j\left(\hat{\bm{x}}_j\right)}.
\end{equation}
A dilated DGF can also be defined recursively.
Specifically, define
\begin{equation}
\label{eq:recursive_dgf}
\begin{aligned}
d(\bm{x}) = & d_{{\myroot}}(\bm{x}_{\myroot}), \\
d_{j}(\bm{x}_{j}) = & \psi_j(\hat{\bm{x}}_j) + \sum_{a \in A_j, j'\in C_{j, a}}{\hat{\bm{x}}_{j}[a] d_{j'}\left(\bm{x}_{j'}\right)}.
\end{aligned}
\end{equation}

\subsection{Recursive Definition of Bregman Divergence}
\label{sec:recursive_BD}
Let's begin by computing the gradient of $d_{j}(\bm{x}_{j})$ for any $j\in \mathcal{J}$.
\begin{lemma}
\label{lm:nabla_d_re}
Given a DGF $d_{j}(\bm{x}_{j}) =  \psi_j(\hat{\bm{x}}_j) + \sum_{a \in A_j, j'\in C_{j, a}}{\hat{\bm{x}}_j[a] d_{j'}\left(\bm{x}_{j'}\right)}$,
\begin{equation}
\nabla_{\bm{x}_{j}[j]} d_j(\bm{x}_{j}) =    \nabla \psi_j(\hat{\bm{x}}_j) + \left(\sum_{j' \in C_{j, a}}{\left(d_{j'}\left(\bm{x}_{j'}\right) - \left \langle  \nabla d_{j'}\left(\bm{x}_{j'}\right), \bm{x}_{j'} \right \rangle\right)}\right)_{a\in A_j},
\end{equation}  
and for any  $a \in A_j$, $j' \in C_{j, a}$,
\begin{equation}
\nabla_{\bm{x}_{j}[\downarrow j']}d_j(\bm{x}_{j}) 
= \nabla d_{j'}\left(\bm{x}_{j'}\right).
\end{equation}
\end{lemma}

\begin{proof}
Let's first calculate the gradient for an action $a \in A_j$:
\begin{equation}
\begin{aligned}
& \nabla_{\bm{x}_{j}[j, a]} d_j(\bm{x}_{j}) =    \nabla_{\bm{x}_{j}[j, a]} \psi_j(\bm{x}_{j}[j]) + \sum_{j'\in C_{j, a}}{\nabla_{\bm{x}_{j}[j, a]} \left(\bm{x}_{j}[j, a] d_{j'}\left(\frac{\bm{x}_{j}[\downarrow j']}{\bm{x}_{j}[j, a]}\right)\right) }.
\end{aligned}
\end{equation}
Since
\begin{equation}
\begin{aligned}
\nabla_{\bm{x}_{j}[j, a]} \left(\bm{x}_{j}[j, a] d_{j'}\left(\frac{\bm{x}_{j}[\downarrow j']}{\bm{x}_{j}[j, a]}\right)\right)
= d_{j'}\left(\frac{\bm{x}_{j}[\downarrow j']}{\bm{x}_{j}[j, a]}\right) - \left \langle  \nabla d_{j'}\left(\frac{\bm{x}_{j}[\downarrow j']}{\bm{x}_{j}[j, a]}\right), \frac{\bm{x}_{j}[\downarrow j']}{\bm{x}_{j}[j, a]} \right \rangle,
\end{aligned}
\end{equation}
we have,
\begin{equation}
\begin{aligned}
& \nabla_{\bm{x}_{j}[j]} d_j(\bm{x}_{j}) =     \nabla \psi_j(\hat{\bm{x}}_j) + \left(\sum_{j' \in C_{j, a}}{\left(d_{j'}\left(\bm{x}_{j'}\right) - \left \langle  \nabla d_{j'}\left(\bm{x}_{j'}\right), \bm{x}_{j'} \right \rangle\right)}\right)_{a\in A_j}.
\end{aligned}
\end{equation}
For $\nabla_{\bm{x}_{j}[\downarrow j']}d_j(\bm{x}_{j})$, we have,
\begin{equation}
\begin{aligned}
&   \nabla_{\bm{x}_{j}[\downarrow j']}d_j(\bm{x}_{j}) 
= \nabla_{\bm{x}_{j}[\downarrow j']} \left(\bm{x}_{j}[j, a] d_{j'}\left(\frac{\bm{x}_{j}[\downarrow j']}{\bm{x}_{j}[j, a]}\right)\right)
= \nabla d_{j'}\left(\frac{\bm{x}_{j}[\downarrow j']}{\bm{x}_{j}[j, a]}\right)
= \nabla d_{j'}\left(\bm{x}_{j'}\right).
\end{aligned}
\end{equation}

\end{proof}

Based on the above lemma, we can construct the Bregman divergence recursively for any dilated DGF.

\begin{lemma}
\label{lm:domp_bregman}
A Bregman divergence $\mathcal{B}_{d}(\bm{x}\|\bm{x}')$ constructed from a dilated DGF $d({\bm{x}})$, which is differentiable and strictly convex on a sequence-form space $\mathcal{X}$, can be constructed recursively as follows:
\begin{equation}
\begin{aligned}
\mathcal{B}_{d}(\bm{x}\|\bm{x}') = & \mathcal{B}_{d_{\myroot}}(\bm{x}_{\myroot}\|\bm{x}'_{\myroot}), \\
\mathcal{B}_{d_j}(\bm{x}_{j}\|\bm{x}'_j) = & \mathcal{B}_{\psi_j}(\hat{\bm{x}}_j \| \hat{\bm{x}}'_j) + \sum_{a\in A_j, j'\in C_{j, a}}{  \hat{\bm{x}}_j[a] \mathcal{B}_{d_{j'}}(\bm{x}_{j'}\| \bm{x}'_{j'})},
\end{aligned}
\end{equation}
where $\mathcal{B}_{\psi_j}(\hat{\bm{x}}_j \| \hat{\bm{x}}'_j)$ is the Bregman divergence constructed from $\psi_{j}(\hat{\bm{x}}_j)$, which is differentiable and strictly convex on $\D^{n_j}$.
\end{lemma}
\begin{proof}

According to Lemma \ref{lm:nabla_d_re}, we have

\begin{equation}
\label{eq:nabla_d_x_x}
\begin{aligned}
& \left\langle \nabla  d_{j}(\bm{x}'_j), \bm{x}_{j} - \bm{x}'_j \right\rangle \\
= &  \left\langle  \nabla \psi_j(\hat{\bm{x}}'_j), \hat{\bm{x}}_j - \hat{\bm{x}}'_j \right\rangle + \sum_{a\in A_j, j'\in C_{j, a}}{ \left( \hat{\bm{x}}_j[a] - \hat{\bm{x}}'_{j}[a] \right) \left( d_{j'}\left(\bm{x}'_{j'}\right) - \left \langle  \nabla d_{j'}\left(\bm{x}'_{j'}\right), \bm{x}'_{j'} \right\rangle \right)} + \\
& \sum_{a \in A_j, j'\in C_{j, a}}{ \left \langle \nabla d_{j'}\left(\bm{x}'_{j'}\right) , \hat{\bm{x}}_j[a] \bm{x}_{j'} - \hat{\bm{x}}'_{j}[a] \bm{x}'_{j'}\right \rangle} \\
= &  \left\langle  \nabla \psi_j(\hat{\bm{x}}'_j), \hat{\bm{x}}_j - \hat{\bm{x}}'_j \right\rangle +  \sum_{a\in A_j, j'\in C_{j, a}}{\left( \hat{\bm{x}}_j[a]d_{j'}\left(\bm{x}'_{j'}\right) - \hat{\bm{x}}'_{j}[a]d_{j'}\left(\bm{x}'_{j'}\right) \right)} + \\
& \sum_{a \in A_j, j'\in C_{j, a}}{ \hat{\bm{x}}_j[a] \left \langle \nabla d_{j'}\left(\bm{x}'_{j'}\right) ,  \bm{x}_{j'} - \bm{x}'_{j'}\right \rangle}
\end{aligned}
\end{equation}

So, the Bregman divergence defined on the dilated DGF is
\begin{equation}
\begin{aligned}
& \mathcal{B}_{d_j}(\bm{x}_{j}\|\bm{x}'_j) = d_{j}(\bm{x}_{j}) -  d_{j}(\bm{x}'_j) - \left\langle \nabla  d_{j}(\bm{x}'_j), \bm{x}_{j} - \bm{x}'_j \right\rangle \\
= &  \psi_j(\bm{x}_{j}[j]) -   \psi_j(\bm{x}'_j[j]) + \sum_{a \in A_j, j'\in C_{j, a}}{\left( \hat{\bm{x}}_j[a] d_{j'}\left(\bm{x}_{j'}\right) -\hat{\bm{x}}'_{j}[a] d_{j'}\left(\bm{x}'_{j'}\right)\right)} -  \left\langle \nabla  d_{j}(\bm{x}'_j), \bm{x}_{j} - \bm{x}'_j \right\rangle \\
= &  \psi_j(\bm{x}_{j}[j]) -   \psi_j(\bm{x}'_j[j]) -  \left\langle  \nabla \psi_j(\hat{\bm{x}}'_j), \hat{\bm{x}}_j - \hat{\bm{x}}'_j \right\rangle + \\
& \sum_{a \in A_j, j'\in C_{j, a}}{\hat{\bm{x}}_j[a]\left( d_{j'}\left(\bm{x}_{j'}\right) - d_{j'}\left(\bm{x}'_{j'}\right) -   \left \langle \nabla d_{j'}\left(\bm{x}'_{j'}\right) ,  \bm{x}_{j'} - \bm{x}'_{j'}\right\rangle \right)} \\
= & \mathcal{B}_{\psi_j}(\hat{\bm{x}}_j \| \hat{\bm{x}}'_j) + \sum_{a\in A_j, j'\in C_{j, a}}{  \hat{\bm{x}}_j[a] \mathcal{B}_{d_{j'}}(\bm{x}_{j'}\| \bm{x}'_{j'})}.
\end{aligned}
\end{equation}
\end{proof}

\subsection{Proof for Proposition \ref{prop:domp_FTRL}}
\begin{proof}

When $q^{0:t}(\bm{x}) = \sum_{j\in \mathcal{J}}{ \bm{x}[p_j]\psi^t_j\left(\hat{\bm{x}}_j\right)}$ is a dilated DGF, according to the recursive definition of dilate DGF, we let 
\begin{equation}
    q^{0:t}_{j}(\bm{x}_{j}) = \sum_{j'\in C_{\downarrow j}}{\bm{x}_j[p_j]\psi^{t}_{j'}\left(\hat{\bm{x}}_{j'}\right)}.
\end{equation}

Define the target function for FTRL as
\begin{equation}
\begin{aligned}
    F^t(\bm{x}) = &\langle \bm{L}^t, \bm{x} \rangle + q^{0:t}(\bm{x}). \\
\end{aligned}
\end{equation}
Then, we have $\bm{x}^{t+1} =  \argmin_{\bm{x}\in \mathcal{X}}{F^t(\bm{x})}$. Furthermore, define
\begin{equation}
\begin{aligned}
F^t_{j}(\bm{x}_{j}) = \langle \bm{L}^t[\downarrow j], \bm{x}_{j} \rangle + q^{0:t}_{j}(\bm{x}_{j}).
\end{aligned}
\end{equation}
Note that $F^t(\bm{x}) = F^t_o(\bm{x}_o)$ and $q^{0:t-1}(\bm{x}) = q^{0:t-1}_o(\bm{x}_o)$.

Since both the sequence-form space and the dilated DGF are defined recursively, we have
\begin{equation}
\begin{aligned}
F^t_{j}(\bm{x}_{j})  =  &  \langle \bm{L}^t[\downarrow j], \bm{x}_{j} \rangle + q^{0:t}_{j}(\bm{x}_{j}) \\
= & \left\{\langle \bm{L}^t[j], \bm{x}_{j}[j] \rangle + \sum_{a\in A_j, j' \in C_{j, a}}{ \langle [\bm{L}^t[\downarrow j'], \bm{x}_{j}[\downarrow j'] \rangle}\right\}  +    \left\{\psi^t_j(\hat{\bm{x}}_{j}) + \sum_{a \in A_j, j'\in C_{j, a}}{\hat{\bm{x}}_j[a] q^{0:t}_{j'}\left(\bm{x}_{j'}\right)}\right\}.
\end{aligned}
\end{equation}
Since ${\bm{x}_{j}[\downarrow j']}/{\bm{x}_{j}[j, a]} = \bm{x}_{j'}$ and $\bm{x}_{j}[j] = \hat{\bm{x}}_{j}$, we have
\begin{equation}
\begin{aligned}
 F^t_{j}(\bm{x}_{j})  
= & \langle \bm{L}^t[j], \hat{\bm{x}}_{j} \rangle  +   \psi^t_j(\hat{\bm{x}}_{j}) + \sum_{a \in A_j, j'\in C_{j, a}}{\hat{\bm{x}}_j[a]  \left( \langle \bm{L}^t[\downarrow j'], \bm{x}_{j'} \rangle +  q^{0:t}_{j'}\left(\bm{x}_{j'}\right)\right)} \\
= & \langle \bm{L}^t[j], \hat{\bm{x}}_{j} \rangle  +   \psi^t_j(\hat{\bm{x}}_{j}) + \sum_{a \in A_j, j'\in C_{j, a}}{\hat{\bm{x}}_j[a] F^t_{j'}(\bm{x}_{j'}) }.
\end{aligned}
\end{equation}
Let $\tilde{\bm{g}}^t_{j} = \left(\sum_{j'\in C_{j, a}}{F^t_{j'}(\bm{x}_{j'}})\right)_{a \in A_j}$,
then 
\begin{equation}
F^t_{j}(\bm{x}_{j}) = \langle \bm{L}^t[j] + \tilde{\bm{g}}^t_{j} , \hat{\bm{x}}_{j} \rangle  +   \psi^t_j(\hat{\bm{x}}_{j}),
\end{equation}
and, 
\begin{equation}
\begin{aligned}
\min_{\bm{x}_{j} \in \mathcal{X}_{j} }{F^t_{j}(\bm{x}_{j})} = &  \min_{\bm{x}_{j} \in \mathcal{X}_{j} }{\left\{\langle \bm{L}^t[j] + \tilde{\bm{g}}^t_{j}, \hat{\bm{x}}_j \rangle  +   \psi^t_j(\hat{\bm{x}}_j)\right\}} \\
= &  \min_{\hat{\bm{x}}_j \in \D^{n_j} }{\left\{\langle \bm{L}^t[j] + \min_{\bm{x}_{j}}{\tilde{\bm{g}}^t_{j}}, \hat{\bm{x}}_j \rangle  +   \psi^t_j(\hat{\bm{x}}_j)\right\}} \\
= &  \min_{\hat{\bm{x}}_j \in \D^{n_j} }{\left\{\langle \bm{L}^t[j] + \hat{\bm{g}}^t_{j}, \hat{\bm{x}}_j \rangle  +   \psi^t_j(\hat{\bm{x}}_j)\right\}} \\
= & - \psi^{*t}_j(-\bm{L}^t[j] - \hat{\bm{g}}^t_{j}).
\end{aligned}
\end{equation}
where $\hat{\bm{g}}^t_{j}[a] = \sum_{j'\in C_{j, a}}{\min_{\hat{\bm{x}}_{j'}\in \mathcal{X}_{j'} } F^{t}_{j'}(\bm{x}_{j'})}$, and 
\begin{equation}
\psi^{*t}_j(-\bm{L}^t[j] - \hat{\bm{g}}^t_{j}) = \max_{\hat{\bm{x}}_j \in \D^{n_j} }{ \left\{\langle - \bm{L}^t[j] - \hat{\bm{g}}^t_{j}, \hat{\bm{x}}_{j} \rangle  -  \psi^t_j(\hat{\bm{x}}_{j})\right\}}.
\end{equation}

So,
\begin{equation}
\hat{\bm{x}}^{t+1}_j =\argmin_{\hat{\bm{x}}_j \in \D^{n_j} }{ \left\{\langle \bm{L}^t[j] + \hat{\bm{g}}^t_{j}, \hat{\bm{x}}_{j} \rangle  +  \psi^t_j(\hat{\bm{x}}_{j})\right\}} = \nabla \psi^{*t}_j(-\bm{L}^t[j] - \hat{\bm{g}}^t_{j}).
\end{equation}

The above equation holds for any $j \in \mathcal{J}$. So, $\hat{\bm{g}}^t_{j}[a] =  \sum_{j'\in C_{j, a}}{\min_{\hat{\bm{x}}_{j'}\in \mathcal{X}_{j'} } F^{t}_{j'}(\bm{x}_{j'})} = \sum_{j'\in C_{j, a}}-\psi^{*t}_{j'}(-\bm{L}^t[j'] - \hat{\bm{g}}^t_{j'})$. In conclusion, 
\begin{equation}
    \hat{\bm{x}}^{t+1}_j = \argmin_{\hat{\bm{x}}_j \in \D^{n_j} }\left\{\left\langle \hat{\bm{L}}'^t_j, \hat{\bm{x}}_j \right\rangle  +  \psi^t_j(\hat{\bm{x}}_j)\right\} = \nabla \psi^{*t}_j(-\hat{\bm{L}}'^t_j),
\end{equation}
where 
\begin{equation}
\hat{\bm{L}}'^t_j[a] = \bm{L}^t[j, a] + \sum_{j'\in C_{j, a}}{-\psi^{*t}_j(-\hat{\bm{L}}'^t_{j'})}.
\end{equation}
\end{proof}

\subsection{Proof for Proposition \ref{prop:domp_OMD}}

\begin{proof}
As in the proof for Proposition \ref{prop:domp_OMD}, we let 
\begin{equation}
    q^{0:t}_{j}(\bm{x}_{j}) = d_{j}(\bm{x}_{j}) = \sum_{j'\in C_{\downarrow j}}{\bm{x}_j[p_j]\psi^{t}_{j'}\left(\hat{\bm{x}}_{j'}\right)}.
\end{equation}
Define the target function for OMD as
\begin{equation}
\begin{aligned}
    G^t(\bm{x}) = \langle \bm{l}^t, \bm{x} \rangle + q^t(\bm{x}) + \mathcal{B}_{q^{0:t-1}}(\bm{x} \| \bm{x}^t).
\end{aligned}
\end{equation}
Then, we have $\bm{x}^{t+1} =  \argmin_{\bm{x}\in \mathcal{X}}{G^t(\bm{x})}$ for OMD. Furthermore, define
\begin{equation}
\begin{aligned}
G^t_{j}(\bm{x}_{j}) = \langle \bm{l}^t[\downarrow j], \bm{x}_j \rangle + q^t_j(\bm{x}_j) + \mathcal{B}_{q^{0:t-1}_j}(\bm{x}_j \| \bm{x}^t_j).
\end{aligned}
\end{equation}
Note that $G^t(\bm{x}) = G^t_o(\bm{x}_o)$.

Since both the sequence-form space and the Bregman divergence are defined recursively, we have
\begin{equation}
\begin{aligned}
& G^t_{j}(\bm{x}_{j})  =  \langle \bm{l}^t[\downarrow j], \bm{x}_{j} \rangle + q^t_j(\bm{x}_{j}) +  \mathcal{B}_{q^{0:t-1}_j}(\bm{x}_{j}\| \bm{x}^t_j) \\
= & \left\{\langle \bm{l}^t[j], \bm{x}_{j}[j] \rangle + \sum_{a\in A_j, j' \in C_{j, a}}{ \langle [\bm{l}^t[\downarrow j'], \bm{x}_{j}[\downarrow j'] \rangle}\right\}  + \\
& \left\{\psi^t_j(\hat{\bm{x}}_j) - \psi^{t-1}_j(\hat{\bm{x}}_j) + \mathcal{B}_{\psi^{t-1}_j}(\hat{\bm{x}}_{j}\|\hat{\bm{x}}^t_{j}) + \sum_{a \in A_j, j'\in C_{j, a}}{\hat{\bm{x}}_j[a]  \left( q^t_{j'}(\bm{x}_{j'}) +  \mathcal{B}_{q^{0:t-1}_{j'}}(\bm{x}_{j'}\| \bm{x}^t_{j'})\right)}\right\} \\
= & \langle \bm{l}^t[j], \hat{\bm{x}}_{j} \rangle +  \psi^t_j(\hat{\bm{x}}_j) - \psi^{t-1}_j(\hat{\bm{x}}_j) +   \mathcal{B}_{\psi^{t-1}_j}(\hat{\bm{x}}_{j}\|\hat{\bm{x}}^t_{j}) + \\
& \sum_{a \in A_j, j'\in C_{j, a}}{\hat{\bm{x}}_j[a]  \left( \langle \bm{l}^t[\downarrow j'], \bm{x}_{j'} \rangle +   q^t_{j'}(\bm{x}_{j'}) + \mathcal{B}_{q^{0:t-1}_{j'}}(\bm{x}_{j'}\| \bm{x}^t_{j'})\right)} \\
= & \langle \bm{l}^t[j], \hat{\bm{x}}_{j} \rangle  +\psi^t_j(\hat{\bm{x}}_j) - \psi^{t-1}_j(\hat{\bm{x}}_j) +   \mathcal{B}_{\psi^{t-1}_j}(\hat{\bm{x}}_{j}\|\hat{\bm{x}}^t_{j}) + \sum_{a \in A_j, j'\in C_{j, a}}{\hat{\bm{x}}_j[a] G^t_{j'}(\bm{x}_{j'}) }.
\end{aligned}
\end{equation}
Let $\tilde{\bm{g}}^t_{j} = \left(\sum_{j'\in C_{j, a}}{G^t_{j'}(\bm{x}_{j'}})\right)_{a \in A_j}$,
then 
\begin{equation}
G^t_{j}(\bm{x}_{j}) = \langle \bm{l}^t[j] + \tilde{\bm{g}}^t_{j} , \hat{\bm{x}}_{j} \rangle + \psi^t_j(\hat{\bm{x}}_j) - \psi^{t-1}_j(\hat{\bm{x}}_j) +   \mathcal{B}_{\psi^{t-1}_j}(\hat{\bm{x}}_{j}\|\hat{\bm{x}}^t_{j}).
\end{equation}

So, 
\begin{equation}
\begin{aligned}
 \min_{\bm{x}_{j} \in \mathcal{X}_{j} }{G^t_{j}(\bm{x}_{j})} = &  \min_{\bm{x}_{j} \in \mathcal{X}_{j} }{\left\{\langle \bm{l}^t[j] + \tilde{\bm{g}}^t_{j}, \hat{\bm{x}}_j \rangle  +   \psi^t_j(\hat{\bm{x}}_j) - \psi^{t-1}_j(\hat{\bm{x}}_j) +   \mathcal{B}_{\psi^{t-1}_j}(\hat{\bm{x}}_{j}\|\hat{\bm{x}}^t_{j})\right\}} \\
= &  \min_{\hat{\bm{x}}_j \in \D^{n_j} }{\left\{\langle \bm{l}^t[j] + \min_{\bm{x}_{j}}{\tilde{\bm{g}}^t_{j}}, \hat{\bm{x}}_j \rangle  +   \psi^t_j(\hat{\bm{x}}_j) - \psi^{t-1}_j(\hat{\bm{x}}_j) +   \mathcal{B}_{\psi^{t-1}_j}(\hat{\bm{x}}_{j}\|\hat{\bm{x}}^t_{j})\right\}} \\
= &  \min_{\hat{\bm{x}}_j \in \D^{n_j} }{\left\{\langle \bm{l}^t[j] + \hat{\bm{g}}^t_{j}, \hat{\bm{x}}_j \rangle  +  \psi^t_j(\hat{\bm{x}}_j) - \psi^{t-1}_j(\hat{\bm{x}}^t_{j}) -  \langle \nabla \psi^{t-1}_j(\hat{\bm{x}}^t_j), \hat{\bm{x}}_j - \hat{\bm{x}}^t_j \rangle\right\}} \\
= & 
{\psi}^{*t-1}_{j}(\nabla \psi^{t-1}_{j}(\hat{\bm{x}}^{t}_{j}))
- \psi^{*t}_j(\nabla \psi^{t-1}_j(\hat{\bm{x}}^t_j) - \bm{l}^t[j] - \hat{\bm{g}}^t_j).
\end{aligned}
\end{equation}
where $\hat{\bm{g}}^t_{j}[a] = \sum_{j'\in C_{j, a}}{\min_{\bm{x}_{j'} \in \mathcal{X}_{j'} }G^{t}_{j'}(\bm{x}_{j'})}$, and 
\begin{equation}
\begin{aligned}
{\psi}^{*t-1}_{j}(\nabla \psi^{t-1}_{j}(\hat{\bm{x}}^{t}_{j})) & \xlongequal{ \text{Fenchel-Young Inequality}}  \langle \nabla \psi^{t-1}_j(\hat{\bm{x}}^t_j), \hat{\bm{x}}^t_j \rangle - \psi^{t-1}_j(\hat{\bm{x}}^t_{j}), \\
\psi^{*t}_j(\nabla \psi^{t-1}_j(\hat{\bm{x}}^t_j) - \bm{l}^t[j] - \hat{\bm{g}}^t_j) & = \max_{\hat{\bm{x}}_j \in \D^{n_j} }{ \left\{\langle \nabla \psi^{t-1}_j(\hat{\bm{x}}^t_j) - \bm{l}^t[j] - \hat{\bm{g}}^t_j, \hat{\bm{x}}_{j} \rangle  - \psi^t_j(\hat{\bm{x}}_{j})\right\}}.
\end{aligned}
\end{equation}
So, 
\begin{equation}
\begin{aligned}
\hat{\bm{x}}^{t+1}_j = & \argmin_{\hat{\bm{x}}_j \in \D^{n_j} }{\left\{\langle \bm{l}^t[j] + \hat{\bm{g}}^t_{j}, \hat{\bm{x}}_j \rangle + \psi^t_j(\hat{\bm{x}}_j) - \psi^{t-1}_j(\hat{\bm{x}}_j) +   \mathcal{B}_{\psi^{t-1}_j}(\hat{\bm{x}}_{j}\|\hat{\bm{x}}^t_{j})\right\}} \\
= & \nabla \psi^{*t}_j(\nabla \psi^{t-1}_j(\hat{\bm{x}}^t_j) - \bm{l}^t[j] - \hat{\bm{g}}^t_j).
\end{aligned}
\end{equation}
The above equation holds for any $j \in \mathcal{J}$. 
In conclusion,
\begin{equation}
\hat{\bm{x}}^{t+1}_j = \nabla \psi^{*t}_{j}(\nabla \psi^{t-1}_{j}(\hat{\bm{x}}^{t}_{j}) -\hat{\bm{l}}'^t_{j}).
\end{equation}
where 
\begin{equation}
\hat{\bm{l}}'^t_j[a] =  \bm{l}^t[j, a] + \sum_{j'\in C_{j, a}}\left\{{\psi}^{*t-1}_{j'}(\nabla \psi^{t-1}_{j'}(\hat{\bm{x}}^{t}_{j'}))-{\psi}^{*t}_{j'}(\nabla \psi^{t-1}_{j'}(\hat{\bm{x}}^{t}_{j'}) -\hat{\bm{l}}'^t_{j'})\right\}.
\end{equation}
\end{proof}

\section{Equivalence Analysis and its Application}
\label{sec:Equivalence}

\subsection{Proof for Theorem \ref{th:fd_ada_FTRL}}

\begin{proof}
According to Lemma 2 and Theorem 3 in \cite{JoulaniGS20}, for both FD-FTRL and FD-OMD, we have 
\begin{equation}
\sum_{t=1}^{T}{\left(\langle \bm l^t, \bm x^t \rangle - \langle \bm l^t, \bm x' \rangle\right)} \leq \sum_{t=0}^{T}\left(q^t(\bm{x}') - q^t(\bm{x}^{t+1})\right) + \sum_{t=1}^{T}\left(\langle \bm{l}^t, \bm{x}^t - \bm{x}^{t+1} \rangle - \mathcal{B}_{q^{0:t-1}}(\bm{x}^{t+1}\|\bm{x}^t) \right). 
\end{equation}
where $q^{0:t-1}(\bm{x}) = \sum_{j\in \mathcal{J}}{\bm{x}[p_j] \beta^{t-1}_j \frac{1}{2} (\|\hat{\bm{x}}_j\|^2_2 + \|\hat{\bm{x}}^{t}_j\|^2_2) }$ and $\mathcal{B}_{q^{0:t-1}}(\bm{x}^{t+1}\|\bm{x}^t) =  \sum_{j\in \mathcal{J}}{\bm{x}^{t+1}[p_{j}] \beta^{t-1}_j \frac{1}{2} \|\hat{\bm{x}}^{t+1}_j - \hat{\bm{x}}^{t}_j \|^2_2 }$. Note that Assumptions 1, 2, 3, 5, and 8 in \cite{JoulaniGS20} have already been fulfilled.

Firstly, for $\sum_{t=0}^{T}q^t(\bm{x}')$, we have
\begin{equation}
\sum_{t=0}^{T}q^t(\bm{x}')= q^{0:T}(\bm{x}') 
=  \sum_{j\in \mathcal{J}}{\bm{x}'[p_{j}] \beta^{T}_j\left(\frac{1}{2}\|\hat{\bm{x}}'_j\|^2_2 + \frac{1}{2}\|\hat{\bm{x}}^{T+1}_j\|^2_2\right)  } 
\leq  \sum_{j\in \mathcal{J}}{\bm{x}'[p_{j}]\beta^{T}_j}.
\end{equation}
For $-q^t(\bm{x}^{t+1})$, we have
\begin{equation}
\begin{aligned}
& - q^t(\bm{x}^{t+1}) =  q^{0:t-1}(\bm{x}^{t+1}) - q^{0:t}(\bm{x}^{t+1})\\
= &  \sum_{j\in \mathcal{J}}{\bm{x}^{t+1}[p_{j}] \beta^{t-1}_j\left(\frac{1}{2}\|\hat{\bm{x}}^{t+1}_j\|^2_2 + \frac{1}{2}\|\hat{\bm{x}}^{t}_j\|^2_2\right)  } - \sum_{j\in \mathcal{J}}{\bm{x}^{t+1}[p_{j}] \beta^t_j\left(\frac{1}{2}\|\hat{\bm{x}}^{t+1}_j\|^2_2 + \frac{1}{2}\|\hat{\bm{x}}^{t+1}_j\|^2_2\right)  } \\
= & \sum_{j\in \mathcal{J}}{ \bm{x}^{t+1}[p_{j}] \left( \frac{1}{2}\beta^{t-1}_j\|\hat{\bm{x}}^{t+1}_j\|^2_2 +   \frac{1}{2}\beta^{t-1}_j\|\hat{\bm{x}}^{t}_j\|^2_2 -  \beta^t_j \|\hat{\bm{x}}^{t+1}_j\|^2_2 \right)}.
\end{aligned}
\end{equation}

On the other hand, 
\begin{equation}
\begin{aligned}
- \mathcal{B}_{q^{0:t-1}}(\bm{x}^{t+1}\|\bm{x}^t) = & - \sum_{j\in \mathcal{J}}{\bm{x}^{t+1}[p_{j}] \beta^{t-1}_j \frac{1}{2} \|\hat{\bm{x}}^{t+1}_j - \hat{\bm{x}}^{t}_j \|^2_2 } \\
= & \sum_{j\in \mathcal{J}}{\bm{x}^{t+1}[p_{j}] \left(-  \frac{1}{2}\beta^{t-1}_j \|\hat{\bm{x}}^{t+1}_j \|^2_2 - \frac{1}{2} \beta^{t-1}_j  \|\hat{\bm{x}}^{t}_j \|^2_2 +  \beta^{t-1}_j \langle \hat{\bm{x}}^{t+1}_j, \hat{\bm{x}}^{t}_j \rangle \right) }.
\end{aligned}
\end{equation}

According to Lemma \ref{lm:surrogate_l} in the paper,
\begin{equation}
\begin{aligned}
\langle \bm{l}^t, \bm{x}^t - \bm{x}^{t+1} \rangle 
= & \sum_{j\in\mathcal{J}}{\bm{x}^{t+1}[p_{j}] \langle \hat{\bm{r}}^t_j, \hat{\bm{x}}^{t+1}_j \rangle }.
\end{aligned}
\end{equation}

Therefore,
\begin{equation}
\begin{aligned}
& - q^t(\bm{x}^{t+1}) + \langle \bm{l}^t, \bm{x}^t - \bm{x}^{t+1} \rangle - \mathcal{B}_{q^{0:t-1}}(\bm{x}^{t+1}\|\bm{x}^t) \\
= & \sum_{j\in \mathcal{J}}{ \bm{x}^{t+1}[p_{j}] \left( \beta^{t-1}_j \langle \hat{\bm{x}}^{t+1}_j, \hat{\bm{x}}^{t}_j \rangle -  \beta^t_j \|\hat{\bm{x}}^{t+1}_j\|^2_2 +  \langle \hat{\bm{r}}^t_j, \hat{\bm{x}}^{t+1}_j\rangle \right)} \\
= & \sum_{j\in \mathcal{J}}{ \bm{x}^{t+1}[p_{j}] \langle \beta^{t-1}_j \hat{\bm{x}}^{t}_j -  \beta^t_j\hat{\bm{x}}^{t+1}_j + \hat{\bm{r}}^t_j, \hat{\bm{x}}^{t+1}_j \rangle }.
\end{aligned}
\end{equation}
According to the Fenchel-Young inequality,  we have
\begin{equation}
\begin{aligned}
\langle \beta^{t-1}_j \hat{\bm{x}}^{t}_j -  \beta^t_j\hat{\bm{x}}^{t+1}_j + \hat{\bm{r}}^t_j, \hat{\bm{x}}^{t+1}_j \rangle \leq & \frac{\|\beta^{t-1}_j \hat{\bm{x}}^{t}_j + \hat{\bm{r}}^t_j\|^2_2}{2\beta^t_j} - \frac{\|\beta^t_j\hat{\bm{x}}^{t+1}_j\|^2_2}{2\beta^t_j} \\
= & \frac{\|\beta^{t-1}_j \hat{\bm{x}}^{t}_j\|^2_2 + \|\hat{\bm{r}}^t_j\|^2_2}{2\beta^t_j} - \frac{\|\beta^t_j\hat{\bm{x}}^{t+1}_j\|^2_2}{2\beta^t_j} \\
= & \frac{\|\hat{\bm{r}}^t_j\|^2_2 + \|\beta^{t-1}_j \hat{\bm{x}}^{t}_j\|^2_2  - \|\beta^t_j\hat{\bm{x}}^{t+1}_j\|^2_2}{2\beta^t_j}.
\end{aligned}
\end{equation}
The first equality is because $ \langle \hat{\bm{r}}^t_j,  \hat{\bm{x}}^{t}_j\rangle = \langle \hat{\bm{l}}^t_{j}, \hat{\bm{x}}^t_{j} \rangle - \langle \hat{\bm{l}}^t_j,  \hat{\bm{x}}^{t}_j\rangle = 0$.
Combining the equations above, we have
\begin{equation}
\label{eq:raw_bound}
\begin{aligned}
\sum_{t=1}^{T}{\left(\langle \bm l^t, \bm x^t \rangle - \langle \bm l^t, \bm x' \rangle\right)} \leq &  \sum_{t=0}^{T}\left(q^t(\bm{x}') - q^t(\bm{x}^{t+1})\right) + \sum_{t=1}^{T}\left(\langle \bm{l}^t, \bm{x}^t - \bm{x}^{t+1} \rangle - \mathcal{B}_{q^{0:t-1}}(\bm{x}^{t+1}\|\bm{x}^t) \right) \\
= & \sum_{j\in \mathcal{J}}{\bm{x}'[p_{j}]\beta^{T}_j} +   \sum_{j\in \mathcal{J}}{\sum_{t=1}^T{ \bm{x}^{t+1}[p_{j}] \langle \beta^{t-1}_j \hat{\bm{x}}^{t}_j -  \beta^t_j\hat{\bm{x}}^{t+1}_j + \hat{\bm{r}}^t_j, \hat{\bm{x}}^{t+1}_j \rangle }} \\
\leq  & \sum_{j\in \mathcal{J}}{\left(\bm{x}'[p_{j}]\beta^{T}_j + \sum_{t=1}^T{ \bm{x}^{t+1}[p_{j}] \frac{\|\hat{\bm{r}}^t_j\|^2_2 + \|\beta^{t-1}_j \hat{\bm{x}}^{t}_j\|^2_2 - \|\beta^t_j\hat{\bm{x}}^{t+1}_j\|^2_2  }{2\beta^{t}_j}}\right)}.
\end{aligned}
\end{equation}
Note that $q^0(\bm{x}) > 0$.
So, 
\begin{equation}
\begin{aligned}
\sum_{t=1}^{T}{\left(\langle \bm l^t, \bm x^t \rangle - \langle \bm l^t, \bm x' \rangle\right)}
\leq   \sum_{j\in \mathcal{J}}{\left(\beta^{T}_j + \sum_{t=1}^T{ \frac{\left[\|\hat{\bm{r}}^t_j\|^2_2 + \|\beta^{t-1}_j \hat{\bm{x}}^{t}_j\|^2_2 - \|\beta^t_j\hat{\bm{x}}^{t+1}_j\|^2_2\right]^+}{2\beta^{t}_j} }\right)}.
\end{aligned}
\end{equation}

\end{proof}

\subsection{Proof for Theorem \ref{th:CFR_eq_FTRL}}

\begin{proof}
We prove the equivalence by recursively proving that the local loss $\hat{\bm{L}}'^t_j$ and the local decision $\hat{\bm{x}}^{t+1}_j$ in FD-FTRL equal $\hat{\bm{L}}^t_j$ and ${[\hat{\bm{R}}^t_j]^+}/{\|[\hat{\bm{R}}^t_j]^+\|_1}$ in CFR-RM, respectively, i.e., the following equations hold at all decision points:
\begin{equation}
\label{eq:FTRL_F_eq_L}
\hat{\bm{L}}'^t_j = \hat{\bm{L}}^t_j,
\end{equation}
\begin{equation}
\label{eq:FTRL_x_eq_R}
\hat{\bm{x}}^{t+1}_j = \frac{[\hat{\bm{R}}^t_j]^+}{\|[\hat{\bm{R}}^t_j]^+\|_1}.
\end{equation}
Let the depth of a decision point $j \in \mathcal{J}$ be the maximum length of the action sequence to the end of the game, denoted by $depth(j)$.
First, at decision point $j\in \mathcal{J}$ that has $depth(j) = 1$,
we have $\hat{\bm{L}}'^t_{j} = \bm{L}^t[j] = \hat{\bm{L}}^t_j$. So (\ref{eq:FTRL_F_eq_L}) holds. Besides, according to Proposition \ref{prop:domp_FTRL} in the paper, we have
\begin{equation}
\begin{aligned}
\hat{\bm{x}}^{t+1}_j = \nabla \psi^{*t}_j(-\hat{\bm{L}}'^t_j) = \nabla \psi^{*t}_j(-\bm{L}^t[j]) = \frac{[\alpha^t_j\bm{1} - \bm{L}^t[j]]^+}{\beta^t_j},
\end{aligned}
\end{equation}
where $\alpha^t_j \in \mathbb{R}$ satisfies $\|[\alpha^t_j\bm{1} -\bm{L}^t[j]]^+\|_1 = \beta^t_j$. Since function $f(\alpha^t_j) \mapsto \|[\alpha^t_j\bm{1} - \bm{L}^t[j]]^+\|_1 $ is monotone ascending and convex in the range of $[\min_a{\bm{L}^t[j, a]}, \infty)$ with the minimum equals zero, $\alpha^t_j$ exists and is unique. Since $\hat{\bm{R}}^t_j  = \sum_{k=1}^t \langle \hat{\bm{l}}^k_{j}, \hat{\bm{x}}^k_{j}\rangle\bm{1} - \hat{\bm{L}}^t_j$, when $\beta^t_j = \|[\hat{\bm{R}}^t_j]^+\|_1$, we can conclude that the unique solution of $\alpha^t_j$ in the equation is $\sum_{k=1}^t \langle \hat{\bm{l}}^k_{j}, \hat{\bm{x}}^k_{j}\rangle$.
So 
\begin{equation}
\label{eq:FTRL_x_eq_R_at_1}
\hat{\bm{x}}^{t+1}_j  = \frac{[\alpha^t_j\bm{1} -\bm{L}^t[j] ]^+}{\beta^t_j} = \frac{[\hat{\bm{R}}^t_j]^+}{\|[\hat{\bm{R}}^t_j]^+\|_1},
\end{equation}
So (\ref{eq:FTRL_x_eq_R}) holds here.

Now, assume that (\ref{eq:FTRL_x_eq_R}) and (\ref{eq:FTRL_F_eq_L}) are satisfied at decision points whose depth is less than or equal to $k$. For a decision point $j\in \mathcal{J}$ that has $depth(j) = k + 1$, we have

\begin{equation}
\hat{\bm{L}}'^t_j[a] = \bm{L}^t[j, a] + \sum_{j'\in C_{j, a}}{-\psi^{*t}_{j'}(-\hat{\bm{L}}'^t_{j'})},
\end{equation}
where
\begin{equation}
\label{eq:FTRL_F_eq_L_at_1}
\begin{aligned}
-\psi^{*t}_{j'}(-\hat{\bm{L}}'^t_{j'}) = & {\langle \hat{\bm{L}}'^t_{j'}, \hat{\bm{x}}^{t+1}_{j'} \rangle +  \psi^t_{j'}(\hat{\bm{x}}^{t+1}_{j'}) }\\
= & \sum_{k=1}^t \langle \hat{\bm{l}}^k_{j'}, \hat{\bm{x}}^k_{j'}\rangle + \left\langle \hat{\bm{L}}^t_{j'} - \sum_{k=1}^t \langle \hat{\bm{l}}^k_{j'}, \hat{\bm{x}}^k_{j'}\rangle, \hat{\bm{x}}^{t+1}_{j'} \right\rangle + \psi^t_{j'}(\hat{\bm{x}}^{t+1}_{j'})  \\
= & \sum_{k=1}^t \langle \hat{\bm{l}}^k_{j'}, \hat{\bm{x}}^k_{j'}\rangle - \beta^t_{j'}\|\hat{\bm{x}}^{t+1}_{j'}\|^2_2 + \frac{1}{2}\beta^t_{j'}\|\hat{\bm{x}}^{t+1}_{j'}\|^2_2 + \frac{1}{2}\beta^t_{j'}\|\hat{\bm{x}}^{t+1}_{j'}\|^2_2 \\
= & \sum_{k=1}^t \langle \hat{\bm{l}}^k_{j'}, \hat{\bm{x}}^k_{j'}\rangle.
\end{aligned}
\end{equation}
So
\begin{equation}
\hat{\bm{L}}'^t_j[a] = \bm{L}^t[j, a] + \sum_{j'\in C_{j, a}}{\sum_{k=1}^t \langle \hat{\bm{l}}^k_{j'}, \hat{\bm{x}}^k_{j'}\rangle},
\end{equation}
which is equal to the cumulative counterfactual loss $\hat{\bm{L}}^t_j[a]$ in CFR-RM,
i.e., (\ref{eq:FTRL_x_eq_R}) holds. Again, we have
\begin{equation}
\label{eq:FTRL_x_at_k_1}
\begin{aligned}
\hat{\bm{x}}^{t+1}_j = \nabla \psi^{*t}_j(-\hat{\bm{L}}'^t_j) = \nabla \psi^{*t}_j(-\hat{\bm{L}}^t_j) = \frac{[\alpha^t_j\bm{1} - \hat{\bm{L}}^t_j]^+}{\beta^t_j},
\end{aligned}
\end{equation}
where $\alpha^t_j$ satisfies $\|[\alpha^t_j\bm{1} -\hat{\bm{L}}^t_j]^+\|_1 = \beta^t_j = \|[\hat{\bm{R}}^t_j]^+\|_1$. Since  $\hat{\bm{R}}^t_j  =  \sum_{k=1}^t \langle \hat{\bm{l}}^k_{j}, \hat{\bm{x}}^k_{j}\rangle\bm{1} - \hat{\bm{L}}^t_j$, we have $\alpha^t_j = \sum_{k=1}^t \langle \hat{\bm{l}}^k_{j}, \hat{\bm{x}}^k_{j}\rangle$, and,
\begin{equation}
\label{eq:FTRL_x_eq_R_at_k_1}
\hat{\bm{x}}^{t+1}_j  = \frac{[\alpha^t_j\bm{1} -\hat{\bm{L}}^t_j ]^+}{\beta^t_j} = \frac{[\hat{\bm{R}}^t_j]^+}{\|[\hat{\bm{R}}^t_j]^+\|_1},
\end{equation}
So, (\ref{eq:FTRL_F_eq_L}) holds. 

Since the local decisions between FD-FTRL and CFR-RM at all decision points are equal, we can conclude that CFR-RM is equivalent to a special case of FD-FTRL with $\beta^t_j = \|[\hat{\bm{R}}^t_j]^+\|_1$ at all decision points.

Now, we prove the equivalence between FD-OMD and CFR-RM+.
We prove the equivalence by recursively proving that the following equations hold at all decision points:
\begin{equation}
\label{eq:OMD_F_eq_l}
\hat{\bm{l}}'^t_j = \hat{\bm{l}}^t_j,
\end{equation}
and
\begin{equation}
\label{eq:OMD_x_eq_q}
\hat{\bm{x}}^{t+1}_j = \frac{\hat{\bm{Q}}^t_j}{\|\hat{\bm{Q}}^t_j\|_1}.
\end{equation}
First, at decision point $j\in \mathcal{J}$ that has $depth(j) = 1$, we have $\hat{\bm{l}}'^t_j = \bm{l}^t[j] = \hat{\bm{l}}^t_j$. So, (\ref{eq:OMD_F_eq_l}) holds. Besides, according to Proposition \ref{prop:domp_OMD}, we have
\begin{equation}
\begin{aligned}
\hat{\bm{x}}^{t+1}_j = \nabla \psi^{*t}_{j}(\nabla \psi^{t-1}_{j}(\hat{\bm{x}}^{t}_{j}) -\hat{\bm{l}}'^t_{j}) = \frac{[\beta^{t-1}_j \hat{\bm{x}}^t_j + \alpha^t_j\bm{1} - \bm{l}^t[j]]^+}{\beta^{t}_j},
\end{aligned}
\end{equation}
where $\alpha^t_j$ fulfills the constraint $\|\hat{\bm{x}}^{t+1}_j\|_1 = 1$, i.e., $\|[\beta^{t-1}_j\hat{\bm{x}}^t_j + \alpha^t_j\bm{1} - \bm{l}^t[j]]^+\|_1 = \beta^{t}_j$. 
Note that $\alpha^t_j$ exists and is unique. Recall that, in CFR-RM+, 
\begin{equation}
    \hat{\bm{Q}}^{t}_j = [\hat{\bm{Q}}^{t-1}_j + \langle \hat{\bm{l}}^t_{j}, \hat{\bm{x}}^t_{j} \rangle -  \hat{\bm{l}}^t_j]^+ = [\|\hat{\bm{Q}}^{t-1}_j\|_1 \hat{\bm{x}}^{t}_j + \langle \hat{\bm{l}}^t_{j}, \hat{\bm{x}}^t_{j} \rangle -  \hat{\bm{l}}^t_j]^+.
\end{equation}
So, when $\beta^{t-1}_j = \|\hat{\bm{Q}}^{t-1}_j\|_1$ and $\beta^{t}_j = \|\hat{\bm{Q}}^{t}_j\|_1$, we can conclude that $\alpha^t_j = \langle \hat{\bm{l}}^t_{j}, \hat{\bm{x}}^t_{j} \rangle$ and
\begin{equation}
\label{eq:OMD_x_eq_R_at_1}
\hat{\bm{x}}^{t+1}_j = \frac{[\beta^{t-1}_j \hat{\bm{x}}^t_j + \alpha^t_j\bm{1} - \bm{l}^t[j]]^+}{\beta^{t}_j} = \frac{\hat{\bm{Q}}^t_{j}}{\|\hat{\bm{Q}}^t_{j}\|_1}.
\end{equation}
So, (\ref{eq:OMD_x_eq_q}) holds.

Now, assume that (\ref{eq:OMD_F_eq_l}) and (\ref{eq:OMD_x_eq_q}) are satisfied at decision points whose depth is less than or equal to $k$. For a decision point $j\in \mathcal{J}$ that has $depth(j) = k + 1$, we have
\begin{equation}
    \hat{\bm{l}}'^t_j[a] = \bm{l}^t[j, a] + \sum_{j'\in C_{j, a}}{\hat{l}'^t_{j'}},
\end{equation}
where
\begin{equation}
\label{eq:OMD_F_eq_l_at_1}
\begin{aligned}
\hat{l}'^t_{j'} = & {\psi}^{*t-1}_{j'}(\nabla \psi^{t-1}_{j'}(\hat{\bm{x}}^{t}_{j'}))-{\psi}^{*t}_{j'}(\nabla \psi^{t-1}_{j'}(\hat{\bm{x}}^{t}_{j'}) -\hat{\bm{l}}'^t_{j'}) \\
= & \left(\langle \beta^{t-1}_{j'} \hat{\bm{x}}^{t}_{j'}, \hat{\bm{x}}^{t}_{j'} \rangle -  \frac{1}{2}\beta^{t-1}_{j'}\|\hat{\bm{x}}^{t}_{j'}\|^2_2 - \frac{1}{2}\beta^{t-1}_{j'}\|\hat{\bm{x}}^{t}_{j'}\|^2_2\right) - \\
& \left(\langle  \beta^{t-1}_{j'} \hat{\bm{x}}^{t}_{j'} - \hat{\bm{l}}^t_{j'}, \hat{\bm{x}}^{t+1}_{j'} \rangle -  \frac{1}{2}\beta^t_{j'}\|\hat{\bm{x}}^{t+1}_{j'}\|^2_2 - \frac{1}{2}\beta^t_{j'}\|\hat{\bm{x}}^{t+1}_{j'}\|^2_2\right) \\
= & \langle \hat{\bm{l}}^t_{j'}, \hat{\bm{x}}^t_{j'}\rangle + \langle \hat{\bm{l}}^t_{j'} - \beta^{t-1}_{j'} \hat{\bm{x}}^t_{j'} - \langle \hat{\bm{l}}^t_{j'}, \hat{\bm{x}}^t_{j'}\rangle, \hat{\bm{x}}^{t+1}_{j'} \rangle +  \beta^t_{j'}\|\hat{\bm{x}}^{t+1}_{j'}\|^2_2\\
= & \langle \hat{\bm{l}}^t_{j'}, \hat{\bm{x}}^t_{j'}\rangle.
\end{aligned}
\end{equation}
So,
\begin{equation}
    \hat{\bm{l}}'^t_j[a] = \bm{l}^t[j, a] + \sum_{j'\in C_{j, a}}{\langle \hat{\bm{l}}^t_{j'}, \hat{\bm{x}}^t_{j'}\rangle},
\end{equation}
which is equal to the counterfactual loss $\hat{\bm{l}}^t_j[a]$ in CFR-RM+, i.e., (\ref{eq:OMD_F_eq_l}) holds. Then, we have
\begin{equation}
\label{eq:OMD_x_at_t_1}
\hat{\bm{x}}^{t+1}_j = \nabla \psi^{*t}_{j}(\nabla \psi^{t-1}_{j}(\hat{\bm{x}}^{t}_{j}) -\hat{\bm{l}}'^t_{j}) = \frac{[\beta^{t-1}_j \hat{\bm{x}}^t_j + \alpha^t_j\bm{1} - \hat{\bm{l}}^t_j]^+}{\beta^{t}_j},
\end{equation}
where $\alpha^t_j$ fulfills $\|[\beta^{t-1}_j \hat{\bm{x}}^t_j + \alpha^t_j\bm{1} - \hat{\bm{l}}^t_j]^+\|_1 = \beta^{t}_j = \|\hat{\bm{Q}}^{t}_j\|_1$. Since  $\hat{\bm{Q}}^{t}_j = [\hat{\bm{Q}}^{t-1}_j + \langle \hat{\bm{l}}^t_{j}, \hat{\bm{x}}^t_{j} \rangle -  \hat{\bm{l}}^t_j]^+ = [\|\hat{\bm{Q}}^{t-1}_j\|_1 \hat{\bm{x}}^t_j + \langle \hat{\bm{l}}^t_{j}, \hat{\bm{x}}^t_{j} \rangle -  \hat{\bm{l}}^t_j]^+$, we have $\alpha^t_j = \langle \hat{\bm{l}}^t_{j}, \hat{\bm{x}}^t_{j} \rangle$ and
\begin{equation}
\label{eq:OMD_x_eq_Q_at_k_1}
\hat{\bm{x}}^{t+1}_j  = \frac{[\beta^{t-1}_j \hat{\bm{x}}^t_j + \alpha^t_j - \hat{\bm{l}}^t_j]^+}{\beta^{t}_j} = \frac{\hat{\bm{Q}}^t_{j}}{\|\hat{\bm{Q}}^t_{j}\|_1},
\end{equation}
So, (\ref{eq:OMD_x_eq_q}) holds.

Since the local decisions between FD-OMD and CFR-RM+ at all decision points are equal, we can conclude that CFR-RM+ is equivalent to a special case of FD-OMD with $\beta^t_j = \|\hat{\bm{Q}}^t_j\|_1$ at all decision points.
\end{proof}

\subsection{Proof for Corollary \ref{co:FD-FTRL(R)_bound}}
We first present a lemma from \cite{FTRL_OMD_proof}. For completeness, the proof is also quoted.
\begin{lemma} 
\label{lm:sum_int}
\cite{FTRL_OMD_proof}
Let $a_0 \geq 0$ and $f: [0, +\infty) \to [0, +\infty)$ a non-increasing function. Then
\begin{equation}
\sum_{t=1}^T{a_t f\left(a_0 + \sum_{k=1}^t a_k\right)} \leq \int_{a_0}^{\sum_{t=0}^Ta_t}f(x)dx.
\end{equation}
\end{lemma}
\begin{proof}
\cite{FTRL_OMD_proof}.
Denote by $s_t = \sum_{k=0}^t a_k$.
\begin{equation}
    a_t f\left(a_0 + \sum_{k=1}^t a_k\right) = a_t f(s_t) \leq \int_{s_{t-1}}^{s_t}f(x)dx.
\end{equation}
Summing over $t =1, \dots, T$, we have the stated bound.
\end{proof}

\begin{proof}
According to Theorem \ref{th:fd_ada_FTRL}, the total regret is
\begin{equation}
\begin{aligned}
R^T \leq & \sum_{j\in \mathcal{J}}{\left(\beta^{T}_j + \sum_{t=1}^T{ \frac{\left[\|\hat{\bm{r}}^t_j\|^2_2 + \lambda^{t-1}_j - \lambda^t_j\right]^+}{2\beta^{t}_j} }\right)} \\
\leq & \sum_{j\in \mathcal{J}}{\left(\beta^{T}_j + \sum_{t=1}^T{ \frac{\|\hat{\bm{r}}^t_j\|^2_2}{2\beta^{t}_j} }\right)} \\
\leq & \sum_{j\in \mathcal{J}}{\sqrt{n_j\lambda^{T}_j}} + \sum_{j\in \mathcal{J}}{\sum_{t=1}^T{ \frac{\|\hat{\bm{r}}^t_j\|^2_2}{2\sqrt{\lambda^t_j}}}}.
\end{aligned}
\end{equation}
The first and the last inequality are because $\beta^t_j = {\sqrt{\lambda^t_j}}/{\|\hat{\bm{x}}^{t+1}_j\|_2}$.
Since $\eta\sum_{k=1}^t{\|\hat{\bm{r}}^k_j\|^2_2} \leq \lambda^t_j$, we have
\begin{equation}
\begin{aligned}
\sum_{t=1}^T{\frac{\|\hat{\bm{r}}^t_j\|^2_2}{2\sqrt{\lambda^t_j}}} \leq & \frac{1}{2\sqrt{\eta}}\sum_{t=1}^T{\frac{\|\hat{\bm{r}}^t_j\|^2_2}{\sqrt{\sum_{k=1}^t{\|\hat{\bm{r}}^k_j\|^2_2}}}} \\
\leq & \frac{1}{\sqrt{\eta}}\sqrt{\sum_{k=1}^T{\|\hat{\bm{r}}^k_j\|^2_2}} \\
\leq & \frac{1}{\eta}\sqrt{\lambda^T_j}.
\end{aligned}
\end{equation}
The second inequality is because of Lemma \ref{lm:sum_int}.
So, 
\begin{equation}
    R^T \leq \sum_{j\in \mathcal{J}}{\left(\sqrt{n_j\lambda^{T}_j} + \frac{1}{\eta}\sqrt{\lambda^T_j}\right)} = \sum_{j\in \mathcal{J}}{\left(\sqrt{n_j} + \frac{1}{\eta}\right)\sqrt{\lambda^T_j}}.
\end{equation}
\end{proof}

\section{Benchmark Games}
\label{sec:descript}

\subsection{Description of the Games}

In this section, we first describe the games we tested in experiments. Then, we measure the games in multiple dimensions.

\begin{itemize}[leftmargin=0.2in]
    \item \textbf{Leduc} \cite{Leduc} is a two-player zero-sum EFG. It can also be considered as a simplified Heads-up Limit Texas Hold'em (HULH)\footnote{https://en.wikipedia.org/wiki/Texas\_hold\_\%27em}. 
    In Leduc, there are two suits of cards with three ranks, and only two rounds of betting are allowed. At the beginning of the game, each player places an ante of one chip in the pot and is dealt with one card, which is only visible to himself. In the first round of betting, player 1 has to choose an action between \textit{Call} and \textit{Raise}. Taking the action Call means that the player will place or has placed the same chips as the opponent and leaves the choice to the opponent. Taking the action Raise means that the player will place more chips than the opponent to the pot. It is only two raises allowed in the first round of betting. Sometimes when a player bets fewer chips than his opponent and is asked to take an action, he can choose to \textit{Fold}. If he does so, then, the game is over, and the player loses all chips. The first round ends if one of the players has chosen Fold or if both players agree to end. If no player folds, a public card is revealed to both of the players and then the second round of betting takes place, with the same dynamic as the first round. After the two rounds of betting, if one of the players has a pair with the public card, that player wins the pot. Otherwise, the player with a higher private card wins. In the first round, the player taking action Raise should place 2 (named raise size) more chips to the pot than the opponent. In the second round, the raise size is 4.
    \item \textbf{Leduc(2, 9)} has the same rules as Leduc, except that it has 2 suits of cards with 9 ranks.
    \item \textbf{FHP(2, 5)} \cite{DeepCFR} is a simplified HULH, and it has 2 suits of cards with 5 ranks. At the beginning of the game, each player places an ante of 50 chips in the pot and is dealt with two cards. 
    It has the same dynamic in each round of betting as Leduc. However, it allows three raises and a raise size of 100 in each round of betting, and the first player to act in the second round is player 2. After the first round of betting, three public cards are revealed to the players. After two rounds of betting, the rank of the 5 cards (2 private cards + 3 public cards) of each player is evaluated, and the player with the higher rank wins the pot. We use the standard hand evaluating method\footnote{ https://en.wikipedia.org/wiki/List\_of\_poker\_hands} used in HULH to evaluate the 5 cards. 
   \item \textbf{Goofspiel} \cite{Goofspiel} is popular benchmark EFG. The game has 3 suits of cards with 5 ranks. At the beginning of the game. Each player is dealt with one suit of private cards. Another suit of cards is served as the prize and kept face down on the desk. At each round of the game, the topmost prize card is revealed. Then the players are asked to select a card from their own cards and reveal the cards simultaneously. The player who has a higher card wins the prize card. If the ranks are equal, the prize card is split. After all prize cards are revealed, the score of each player is the sum of the rank of the prize cards he won. Then, the payoff for the player with the higher score is $+1$, and the payoff for the opponent is $-1$. If the scores are the same, the payoffs for the players are $(0.5, 0.5)$.
    \item \textbf{Goofspiel 4} is the same as Goofspiel, except that Goofspiel 4 has 3 suits of cards with 4 ranks.
    \item \textbf{Goofspiel 4 (imp)} Goofspiel 4 (imp), i.e., Goofspiel 4 with imperfect information, is a variant of Goofspiel 4, which is first proposed in \cite{MCCFR}. In each round, the cards that the players selected are not revealed to each other. Instead, a coordinator will see the cards and determine which player wins the prize card. 
    \item \textbf{Liar's dice} \cite{Liars} is another popular benchmark EFG. At the beginning of the game. Each player secretly rolls a dice. Then the first player claims the outcome of their roll, in the form of \textbf{Q}uantity-\textbf{V}alue, e.g., a claim of $1$-$2$ means that the player believes that there is \textit{one} dice with a face value of \textit{2}. The second player can make a higher claim: either to claim a higher Q with any V or to claim the same Q with a higher V. Otherwise, he can call his opponent a ``liar''. When the player calls a liar, the dice are revealed. If the opponent's claim is false, the player that called a liar on the opponent wins the game; otherwise, the player loses the game.
    \item \textbf{Battleship} \cite{Battleship} is a classic board game. At the beginning of the game, each player secretly places a set of ships on separate grids without overlapping. In this setting, the size of the ship is $1\times 2$ and the value is 4. The size of the grid for each player is $3\times 2$. After all the ships are placed. The players take turns to fire the opponent's ship. Ships that have been hit are considered sunk. The game ends if a player's all ships are all sunk or each of the players has completed 3 shots. Finally, the payoff of the player is calculated as the sum of the values of the opponent's ships that were sunk, minus the sum of the values of the player's ships that were lost.
    
\end{itemize}

\subsection{Measuring the games}

To better measure the games, another model known as the \textit{history tree} is used to describe the games. 
For a two-player imperfect information EFG, we have two players $P=\{1,2\}$. A special player, the \textit{chance} player $c$, is introduced to take action for random events in the game. A history $h\in H$ (i.e., a full information state) is represented as a string of actions that taken by all the players and the chance player. The histories of an EFG form a history tree by nature. Given a non-terminal history $h\in H$, $A(h) \subseteq \mathcal{A}$ gives the set of legal actions, and $P(h)\in P\cup \{c\}$ gives the player that should act in the history. If action $a\in A(h)$ leads from $h$ to $h'$, then $h'=ha$. Denote the set of histories that are reached immediately after taking any action $a\in A(h)$ for $h$ as $C_h = \{h'| h'= ha, a \in A(h)\}$.
In an imperfect information game, histories of player $p \in P$ that are indistinguishable are collected into an \textit{information set} (infoset) $I\in \mathcal{I}_p$. According to the definition of decision points, we know that infosets are actually decision points. So, $\mathcal{J}_p = \mathcal{I}_p$ for any player $p \in P$.

With the above definitions, we can measure the game in multiple dimensions.
In Table \ref{tab:head}, we give some data about the games. In the table, \#Histories measures the number of histories of the game.  Depth measures the depth of the history tree of the game, i.e., the maximum length of the histories. \#Leaves measures the number of leaves of a history tree.
The size of decisions measures the maximum number of histories that belong to the same decision point (infoset). The action factor, denoted by $m$, measures the contribution of the  players to the number of leaves. We define $m$ recursively: if history $h$ belongs to $p\in P$, then $m_h = \sum_{h' \in C_h}{m_{h'}}$; otherwise, $m_h = \max_{h'\in C_h}{m_{h'}}$. Then, the action factor $m$ is defined as $m=m_o$. The stochastic factor, denoted by $s$, measures the contribution of the chance player to the number of leaves. The stochastic factor is defined as follows: if history $h$ belongs to $p\in P$, $s_h = \max_{h' \in C_h}{s_{h'}}$; otherwise, $s_h = \sum_{h'\in C_h}{s_{h'}}$. Then, the stochastic factor $s$ is defined as $s = s_o$.

\begin{table}[htbp]
	\centering
	\caption{Dimensions of the games}\label{tab:head} 
	\begin{tabular}{lcccccccc}
		\toprule
		\multirow{ 2}{*}{Game} & \#Decision & \multirow{ 2}{*}{\#Histories} & \multirow{ 2}{*}{\#Leaves} & \multirow{ 2}{*}{Depth} & size of   & Action            & Stochastic        \\
		                       & points                       &                               &                            &                         & decisions & factor            & factor            \\
		\midrule
		Leduc                  & 936                          & 3780                          & 5520                       & 8                       & 5         & 49                & 120               \\
		Leduc(2, 9)            & 9288                         & $1.5 \times 10^5$             & $2.2 \times 10^5$          & 8                       & 17        & 49                & 4896              \\
		Goof. 4            & 7304                         & $1.1 \times 10^4$             & $1.4 \times 10^4$          & 6                       & 4         & 576               & 24                \\
		Goof. 4 (imp)      & 3608                         & $1.1 \times 10^4$             & $1.4 \times 10^4$          & 6                       & 14        & 576               & 24                \\
		FHP(2, 5)              & $1.4 \times 10^5$            & $1.4 \times 10^6$             & $2.3 \times 10^6$          & 12                      & 9         & 98                & $2.5 \times 10^4$ \\
		Goofspiel              & $9.1 \times 10^5$            & $1.4 \times 10^6$             & $1.7 \times 10^6$          & 8                       & 5         & $1.4 \times 10^4$ & 36                \\
		Liar's dice            & $2.5 \times 10^4$            & $1.5 \times 10^5$             & $1.5 \times 10^5$          & 13                      & 6         & 4095              & 24                \\
		Battleship             & $4.1 \times 10^5$            & $2.3 \times 10^6$             & $7.0 \times 10^6$          & 10                      & 22        & $7.0 \times 10^6$ & 1                 \\
		\bottomrule
	\end{tabular}
\end{table}

\section{Additional Results}
\label{sec:results}

\subsection{Full Results in Eight Games}

\begin{figure}[htbp]
	\centering
	\begin{subfigure}[b]{0.28\linewidth}
        \centering
        \resizebox{\linewidth}{!}{\input{images/plotleduc_poker_ave_compare_iter_0.pgf}}
        \vspace*{-0.25in}

        \caption{Leduc}
    \end{subfigure}
    \hspace*{0.001\linewidth}
    \begin{subfigure}[b]{0.28\linewidth}
        \centering
        \resizebox{\linewidth}{!}{\input{images/plotleduc18_poker_ave_compare_iter_0.pgf}}
        \vspace*{-0.25in}

        \caption{Leduc(2, 9)}
    \end{subfigure}
    \hspace*{0.001\linewidth}
    \begin{subfigure}[b]{0.28\linewidth}
        \centering
        \resizebox{\linewidth}{!}{\input{images/plotgoofspiel_4_ave_compare_iter_0.pgf}}
        \vspace*{-0.25in}

        \caption{Goofspiel 4}
    \end{subfigure}
    \\
    \begin{subfigure}[b]{0.28\linewidth}
        \centering
        \resizebox{\linewidth}{!}{\input{images/plotgoofspiel_4_imp_ave_compare_iter_0.pgf}}
        \vspace*{-0.25in}

        \caption{Goofspiel 4(imp.)}
    \end{subfigure}
    \hspace*{0.001\linewidth}
    \begin{subfigure}[b]{0.28\linewidth}
        \centering
        \resizebox{\linewidth}{!}{\input{images/plotFHP2_poker_ave_compare_iter_0.pgf}}
        \vspace*{-0.25in}

        \caption{FHP(2, 5)}
    \end{subfigure}
    \hspace*{0.001\linewidth}
    \begin{subfigure}[b]{0.28\linewidth}
        \centering
        \resizebox{\linewidth}{!}{\input{images/plotgoofspiel_5_ave_compare_iter_0.pgf}}
        \vspace*{-0.25in}

        \caption{Goofspiel}
    \end{subfigure}
    \\
    \begin{subfigure}[b]{0.28\linewidth}
        \centering
        \resizebox{\linewidth}{!}{\input{images/plotliars_dice_ave_compare_iter_0.pgf}}
        \vspace*{-0.25in}

        \caption{Liar's dice}
    \end{subfigure}
    \hspace*{0.001\linewidth}
    \begin{subfigure}[b]{0.28\linewidth}
        \centering
        \resizebox{\linewidth}{!}{\input{images/plotbattleship_3_2_3_ave_compare_iter_0.pgf}}
        \vspace*{-0.25in}

        \caption{Battleship}
    \end{subfigure}
    \hspace*{0.001\linewidth}
    \begin{subfigure}[b]{0.28\linewidth}
        \includegraphics[height=0.8in]{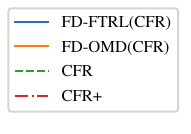}
        \vspace*{0.5in}
    \end{subfigure}
    \caption{Exploitability curves of FD-FTRL(CFR), FD-OMD(CFR), and CFRs in eight games.}
    \label{fig:comp_FTRL}
\end{figure}

In Figure \ref{fig:comp_FTRL}, we compare FD-FTRL(CFR), FD-OMD(CFR) with CFRs in eight games. As we can see, FD-FTRL(CFR) recovers vanilla CFR and FD-OMD(CFR) recovers CFR+. Note that both FD-OMD(CFR) and CFR+ use \textbf{LA} for computing the average strategy.

\begin{figure}[htbp]
	\centering
    \begin{subfigure}[b]{0.28\linewidth}
        \centering
        \resizebox{\linewidth}{!}{\input{images/plotleduc_poker_ave_compare_iter_3.pgf}}
        \vspace*{-0.25in}
        \caption{Leduc}
    \end{subfigure}
    \hspace*{0.001\linewidth}
    \begin{subfigure}[b]{0.28\linewidth}
        \centering
        \resizebox{\linewidth}{!}{\input{images/plotleduc18_poker_ave_compare_iter_3.pgf}}
        \vspace*{-0.25in}
        \caption{Leduc(2, 9)}
    \end{subfigure}
    \hspace*{0.001\linewidth}
    \begin{subfigure}[b]{0.28\linewidth}
        \centering
        \resizebox{\linewidth}{!}{\input{images/plotgoofspiel_4_ave_compare_iter_3.pgf}}
        \vspace*{-0.25in}
        \caption{Goofspiel 4}
    \end{subfigure}
    \\
    \begin{subfigure}[b]{0.28\linewidth}
        \centering
        \resizebox{\linewidth}{!}{\input{images/plotgoofspiel_4_imp_ave_compare_iter_3.pgf}}
        \vspace*{-0.25in}
        \caption{Goofspiel 4(imp.)}
    \end{subfigure}
    \hspace*{0.001\linewidth}
    \begin{subfigure}[b]{0.28\linewidth}
        \centering
        \resizebox{\linewidth}{!}{\input{images/plotFHP2_poker_ave_compare_iter_3.pgf}}
        \vspace*{-0.25in}
        \caption{FHP(2, 5)}
    \end{subfigure}
    \hspace*{0.001\linewidth}
    \begin{subfigure}[b]{0.28\linewidth}
        \centering
        \resizebox{\linewidth}{!}{\input{images/plotgoofspiel_5_ave_compare_iter_3.pgf}}
        \vspace*{-0.25in}
        \caption{Goofspiel}
    \end{subfigure}
    \\
    \begin{subfigure}[b]{0.28\linewidth}
        \centering
        \resizebox{\linewidth}{!}{\input{images/plotliars_dice_ave_compare_iter_3.pgf}}
        \vspace*{-0.25in}
        \caption{Liar's dice}
    \end{subfigure}
    \hspace*{0.001\linewidth}
    \begin{subfigure}[b]{0.28\linewidth}
        \centering
        \resizebox{\linewidth}{!}{\input{images/plotbattleship_3_2_3_ave_compare_iter_3.pgf}}
        \vspace*{-0.25in}
        \caption{Battleship}
    \end{subfigure}
    \hspace*{0.001\linewidth}
    \begin{subfigure}[b]{0.28\linewidth}
        \includegraphics[height=1.0in]{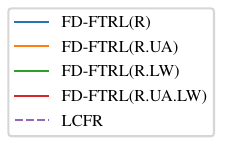}
        \vspace*{0.3in}
    \end{subfigure}
    \caption{Exploitability curves of FD-FTRL(R) in different configurations in eight games.}
    \label{fig:comp_FD-FTRL(R)}
\end{figure}

\begin{figure}[htbp]
	\centering
    \begin{subfigure}[b]{0.28\linewidth}
        \centering
        \resizebox{\linewidth}{!}{\input{images/plotleduc_poker_ave_compare_iter_4.pgf}}
        \vspace*{-0.25in}

        \caption{Leduc}
    \end{subfigure}
    \hspace*{0.001\linewidth}
    \begin{subfigure}[b]{0.28\linewidth}
        \centering
        \resizebox{\linewidth}{!}{\input{images/plotleduc18_poker_ave_compare_iter_4.pgf}}
        \vspace*{-0.25in}

        \caption{Leduc(2, 9)}
    \end{subfigure}
    \hspace*{0.001\linewidth}
    \begin{subfigure}[b]{0.28\linewidth}
        \centering
        \resizebox{\linewidth}{!}{\input{images/plotgoofspiel_4_ave_compare_iter_4.pgf}}
        \vspace*{-0.25in}

        \caption{Goofspiel 4}
    \end{subfigure}
    \\
    \begin{subfigure}[b]{0.28\linewidth}
        \centering
        \resizebox{\linewidth}{!}{\input{images/plotgoofspiel_4_imp_ave_compare_iter_4.pgf}}
        \vspace*{-0.25in}

        \caption{Goofspiel 4(imp.)}
    \end{subfigure}
    \hspace*{0.001\linewidth}
    \begin{subfigure}[b]{0.28\linewidth}
        \centering
        \resizebox{\linewidth}{!}{\input{images/plotFHP2_poker_ave_compare_iter_4.pgf}}
        \vspace*{-0.25in}

        \caption{FHP(2, 5)}
    \end{subfigure}
    \hspace*{0.001\linewidth}
    \begin{subfigure}[b]{0.28\linewidth}
        \centering
        \resizebox{\linewidth}{!}{\input{images/plotgoofspiel_5_ave_compare_iter_4.pgf}}
        \vspace*{-0.25in}

        \caption{Goofspiel}
    \end{subfigure}
    \\
    \begin{subfigure}[b]{0.28\linewidth}
        \centering
        \resizebox{\linewidth}{!}{\input{images/plotliars_dice_ave_compare_iter_4.pgf}}
        \vspace*{-0.25in}

        \caption{Liar's dice}
    \end{subfigure}
    \hspace*{0.001\linewidth}
    \begin{subfigure}[b]{0.28\linewidth}
        \centering
        \resizebox{\linewidth}{!}{\input{images/plotbattleship_3_2_3_ave_compare_iter_4.pgf}}
        \vspace*{-0.25in}

        \caption{Battleship}
    \end{subfigure}
    \hspace*{0.001\linewidth}
    \begin{subfigure}[b]{0.28\linewidth}
        \includegraphics[height=1.0in]{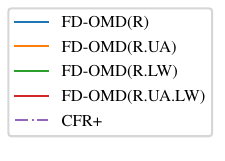}
        \vspace*{0.3in}
    \end{subfigure}
    \caption{Exploitability curves of FD-OMD(R) in different configurations in eight games.}
    \label{fig:comp_FD-OMD(R)}
\end{figure}

\begin{figure}[htbp]
	\centering
    \begin{subfigure}[b]{0.28\linewidth}
        \centering
        \resizebox{\linewidth}{!}{\input{images/plotleduc_poker_ave_compare_iter_1.pgf}}
        \vspace*{-0.25in}
        \caption{Leduc}
    \end{subfigure}
    \hspace*{0.001\linewidth}
    \begin{subfigure}[b]{0.28\linewidth}
        \centering
        \resizebox{\linewidth}{!}{\input{images/plotleduc18_poker_ave_compare_iter_1.pgf}}
        \vspace*{-0.25in}
        \caption{Leduc(2, 9)}
    \end{subfigure}
    \hspace*{0.001\linewidth}
    \begin{subfigure}[b]{0.28\linewidth}
        \centering
        \resizebox{\linewidth}{!}{\input{images/plotgoofspiel_4_ave_compare_iter_1.pgf}}
        \vspace*{-0.25in}
        \caption{Goofspiel 4}
    \end{subfigure}
    \\
    \begin{subfigure}[b]{0.28\linewidth}
        \centering
        \resizebox{\linewidth}{!}{\input{images/plotgoofspiel_4_imp_ave_compare_iter_1.pgf}}
        \vspace*{-0.25in}
        \caption{Goofspiel 4(imp.)}
    \end{subfigure}
    \hspace*{0.001\linewidth}
    \begin{subfigure}[b]{0.28\linewidth}
        \centering
        \resizebox{\linewidth}{!}{\input{images/plotFHP2_poker_ave_compare_iter_1.pgf}}
        \vspace*{-0.25in}
        \caption{FHP(2, 5)}
    \end{subfigure}
    \hspace*{0.001\linewidth}
    \begin{subfigure}[b]{0.28\linewidth}
        \centering
        \resizebox{\linewidth}{!}{\input{images/plotgoofspiel_5_ave_compare_iter_1.pgf}}
        \vspace*{-0.25in}
        \caption{Goofspiel}
    \end{subfigure}
    \\
    \begin{subfigure}[b]{0.28\linewidth}
        \centering
        \resizebox{\linewidth}{!}{\input{images/plotliars_dice_ave_compare_iter_1.pgf}}
        \vspace*{-0.25in}
        \caption{Liar's dice}
    \end{subfigure}
    \hspace*{0.001\linewidth}
    \begin{subfigure}[b]{0.28\linewidth}
        \centering
        \resizebox{\linewidth}{!}{\input{images/plotbattleship_3_2_3_ave_compare_iter_1.pgf}}
        \vspace*{-0.25in}
        \caption{Battleship}
    \end{subfigure}
    \hspace*{0.001\linewidth}
    \begin{subfigure}[b]{0.28\linewidth}
        \includegraphics[height=0.8in]{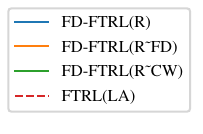}
        \vspace*{0.5in}
    \end{subfigure}
    \caption{Exploitability curves of FD-FTRL(R) with or without certain components in eight games.}
    \label{fig:ablation_FTRL}
\end{figure}

\begin{figure}[htbp]
	\centering
    \begin{subfigure}[b]{0.28\linewidth}
        \centering
        \resizebox{\linewidth}{!}{\input{images/plotleduc_poker_ave_compare_iter_2.pgf}}
        \vspace*{-0.25in}
        \caption{Leduc}
    \end{subfigure}
    \hspace*{0.001\linewidth}
    \begin{subfigure}[b]{0.28\linewidth}
        \centering
        \resizebox{\linewidth}{!}{\input{images/plotleduc18_poker_ave_compare_iter_2.pgf}}
        \vspace*{-0.25in}
        \caption{Leduc(2, 9)}
    \end{subfigure}
    \hspace*{0.001\linewidth}
    \begin{subfigure}[b]{0.28\linewidth}
        \centering
        \resizebox{\linewidth}{!}{\input{images/plotgoofspiel_4_ave_compare_iter_2.pgf}}
        \vspace*{-0.25in}
        \caption{Goofspiel 4}
    \end{subfigure}
    \\
    \begin{subfigure}[b]{0.28\linewidth}
        \centering
        \resizebox{\linewidth}{!}{\input{images/plotgoofspiel_4_imp_ave_compare_iter_2.pgf}}
        \vspace*{-0.25in}
        \caption{Goofspiel 4(imp.)}
    \end{subfigure}
    \hspace*{0.001\linewidth}
    \begin{subfigure}[b]{0.28\linewidth}
        \centering
        \resizebox{\linewidth}{!}{\input{images/plotFHP2_poker_ave_compare_iter_2.pgf}}
        \vspace*{-0.25in}
        \caption{FHP(2, 5)}
    \end{subfigure}
    \hspace*{0.001\linewidth}
    \begin{subfigure}[b]{0.28\linewidth}
        \centering
        \resizebox{\linewidth}{!}{\input{images/plotgoofspiel_5_ave_compare_iter_2.pgf}}
        \vspace*{-0.25in}
        \caption{Goofspiel}
    \end{subfigure}
    \\
    \begin{subfigure}[b]{0.28\linewidth}
        \centering
        \resizebox{\linewidth}{!}{\input{images/plotliars_dice_ave_compare_iter_2.pgf}}
        \vspace*{-0.25in}
        \caption{Liar's dice}
    \end{subfigure}
    \hspace*{0.001\linewidth}
    \begin{subfigure}[b]{0.28\linewidth}
        \centering
        \resizebox{\linewidth}{!}{\input{images/plotbattleship_3_2_3_ave_compare_iter_2.pgf}}
        \vspace*{-0.25in}
        \caption{Battleship}
    \end{subfigure}
    \hspace*{0.001\linewidth}
    \begin{subfigure}[b]{0.28\linewidth}
        \includegraphics[height=0.8in]{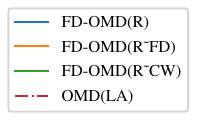}
        \vspace*{0.5in}
    \end{subfigure}
    \caption{Exploitability curves of FD-OMD(R) with or without certain components in eight games.}
    \label{fig:ablation_OMD}
\end{figure}

In Figure \ref{fig:comp_FD-FTRL(R)}, we compare FD-FTRL(R) in different configurations in eight games. As we have stated in the paper, the default FD-FTRL(R) (with \textbf{LA} and \textbf{CW}) has the fastest convergence rate. In Figure \ref{fig:comp_FD-OMD(R)}, we compare FD-OMD(R) in different configurations in eight games. Similar to the conclusion for FD-FTRL(R), the default FD-OMD(R) is the fastest, it is even faster than CFR+ in four games. In Figure \ref{fig:ablation_FTRL} and \ref{fig:ablation_OMD}, we perform an ablation study for FD-FTRL(R) and FD-OMD(R), respectively. As we have stated in the paper, \textbf{CW} has a significant impact on performance. Sometimes, FD-FTRL(R\char126FD) is faster than the default FD-FTRL(R). However, FD-OMD(R) is always faster than the other variants.

\begin{figure}[htbp]
	\centering
    \begin{subfigure}[b]{0.28\linewidth}
        \centering
        \resizebox{\linewidth}{!}{\input{images/plotleduc_poker_ave_compare_iter_5.pgf}}
        \vspace*{-0.25in}

        \caption{Leduc}
    \end{subfigure}
    \hspace*{0.001\linewidth}
    \begin{subfigure}[b]{0.28\linewidth}
        \centering
        \resizebox{\linewidth}{!}{\input{images/plotleduc18_poker_ave_compare_iter_5.pgf}}
        \vspace*{-0.25in}

        \caption{Leduc(2, 9)}
    \end{subfigure}
    \hspace*{0.001\linewidth}
    \begin{subfigure}[b]{0.28\linewidth}
        \centering
        \resizebox{\linewidth}{!}{\input{images/plotgoofspiel_4_ave_compare_iter_5.pgf}}
        \vspace*{-0.25in}

        \caption{Goofspiel 4}
    \end{subfigure}
    \\
    \begin{subfigure}[b]{0.28\linewidth}
        \centering
        \resizebox{\linewidth}{!}{\input{images/plotgoofspiel_4_imp_ave_compare_iter_5.pgf}}
        \vspace*{-0.25in}

        \caption{Goofspiel 4(imp.)}
    \end{subfigure}
    \hspace*{0.001\linewidth}
    \begin{subfigure}[b]{0.28\linewidth}
        \centering
        \resizebox{\linewidth}{!}{\input{images/plotFHP2_poker_ave_compare_iter_5.pgf}}
        \vspace*{-0.25in}

        \caption{FHP(2, 5)}
    \end{subfigure}
    \hspace*{0.001\linewidth}
    \begin{subfigure}[b]{0.28\linewidth}
        \centering
        \resizebox{\linewidth}{!}{\input{images/plotgoofspiel_5_ave_compare_iter_5.pgf}}
        \vspace*{-0.25in}

        \caption{Goofspiel}
    \end{subfigure}
    \\
    \begin{subfigure}[b]{0.28\linewidth}
        \centering
        \resizebox{\linewidth}{!}{\input{images/plotliars_dice_ave_compare_iter_5.pgf}}
        \vspace*{-0.25in}

        \caption{Liar's dice}
    \end{subfigure}
    \hspace*{0.001\linewidth}
    \begin{subfigure}[b]{0.28\linewidth}
        \centering
        \resizebox{\linewidth}{!}{\input{images/plotbattleship_3_2_3_ave_compare_iter_5.pgf}}
        \vspace*{-0.25in}

        \caption{Battleship}
    \end{subfigure}
    \hspace*{0.001\linewidth}
    \begin{subfigure}[b]{0.28\linewidth}
        \includegraphics[height=0.8in]{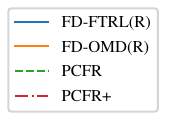}
        \vspace*{0.5in}
    \end{subfigure}
    \caption{Exploitability curves of FD-FTRL(R), FD-OMD(R), PCFR, and PCFR+ in eight games.}
    \label{fig:comp_PCFR}
\end{figure}

In Figure \ref{fig:comp_PCFR}, we compare FD-FTRL(R) and FD-OMD(R) with Predicted CFR (PCFR) and Predicted CFR+ (PCFR+) \cite{Blackwell}. We implement PCFR with \textbf{UA} and implement PCFR+ with \textbf{LA}. As we can see, FD-OMD(R) is still competitive compared to PCFR+.

\begin{figure}[htbp]
	\centering
    \begin{subfigure}[b]{0.28\linewidth}
        \centering
        \resizebox{\linewidth}{!}{\input{images/plotleduc_poker_ave_compare_iter_7.pgf}}
        \vspace*{-0.25in}

        \caption{Leduc}
    \end{subfigure}
    \hspace*{0.001\linewidth}
    \begin{subfigure}[b]{0.28\linewidth}
        \centering
        \resizebox{\linewidth}{!}{\input{images/plotleduc18_poker_ave_compare_iter_7.pgf}}
        \vspace*{-0.25in}

        \caption{Leduc(2, 9)}
    \end{subfigure}
    \hspace*{0.001\linewidth}
    \begin{subfigure}[b]{0.28\linewidth}
        \centering
        \resizebox{\linewidth}{!}{\input{images/plotgoofspiel_4_ave_compare_iter_7.pgf}}
        \vspace*{-0.25in}

        \caption{Goofspiel 4}
    \end{subfigure}
    \\
    \begin{subfigure}[b]{0.28\linewidth}
        \centering
        \resizebox{\linewidth}{!}{\input{images/plotgoofspiel_4_imp_ave_compare_iter_7.pgf}}
        \vspace*{-0.25in}

        \caption{Goofspiel 4(imp.)}
    \end{subfigure}
    \hspace*{0.001\linewidth}
    \begin{subfigure}[b]{0.28\linewidth}
        \centering
        \resizebox{\linewidth}{!}{\input{images/plotFHP2_poker_ave_compare_iter_7.pgf}}
        \vspace*{-0.25in}

        \caption{FHP(2, 5)}
    \end{subfigure}
    \hspace*{0.001\linewidth}
    \begin{subfigure}[b]{0.28\linewidth}
        \centering
        \resizebox{\linewidth}{!}{\input{images/plotgoofspiel_5_ave_compare_iter_7.pgf}}
        \vspace*{-0.25in}

        \caption{Goofspiel}
    \end{subfigure}
    \\
    \begin{subfigure}[b]{0.28\linewidth}
        \centering
        \resizebox{\linewidth}{!}{\input{images/plotliars_dice_ave_compare_iter_7.pgf}}
        \vspace*{-0.25in}

        \caption{Liar's dice}
    \end{subfigure}
    \hspace*{0.001\linewidth}
    \begin{subfigure}[b]{0.28\linewidth}
        \centering
        \resizebox{\linewidth}{!}{\input{images/plotbattleship_3_2_3_ave_compare_iter_7.pgf}}
        \vspace*{-0.25in}

        \caption{Battleship}
    \end{subfigure}
    \hspace*{0.001\linewidth}
    \begin{subfigure}[b]{0.28\linewidth}
        \includegraphics[height=1.2in]{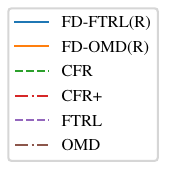}
        \vspace*{0.1in}
    \end{subfigure}
    \caption{Exploitability curves of FD-FTRL(R), FD-OMD(R), FTRL, OMD, and CFRs in eight games.}
    \label{fig:comp_LCFR}
\end{figure}

In Figure \ref{fig:comp_LCFR}, the results of conventional FTRL, OMD, and vanilla CFR that use \textbf{UA} are given. As we can see, both FD-FTRL(R) and FD-OMD(R) are always faster than them.

\subsection{Hyper-parameter Tuning}
\begin{figure}[htbp]
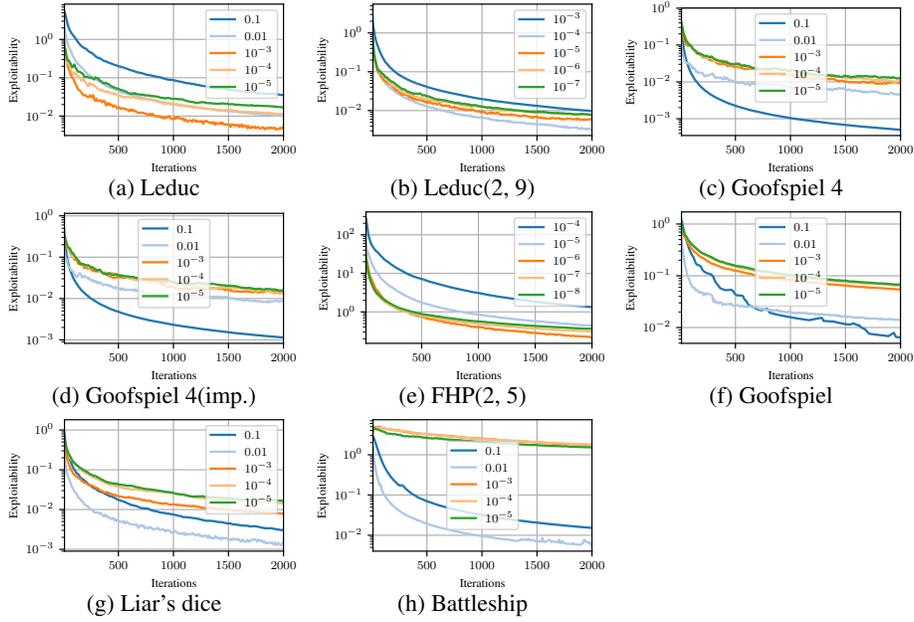

	\centering
    \begin{subfigure}[b]{0.28\linewidth}
        \centering
        \resizebox{\linewidth}{!}{\input{images/plotleduc_poker_sbcfr_olo.sh_game_leduc_poker_cfr_PostSbCFR_a_LinearOpponent_w_Constant_rm_0.1_iter_6.pgf}}
        \vspace*{-0.25in}

        \caption{Leduc}
    \end{subfigure}
    \hspace*{0.001\linewidth}
    \begin{subfigure}[b]{0.28\linewidth}
        \centering
        \resizebox{\linewidth}{!}{\input{images/plotleduc18_poker_sbcfr_olo.sh_game_leduc18_poker_cfr_PostSbCFR_a_LinearOpponent_w_Constant_rm_0.001_iter_6.pgf}}
        \vspace*{-0.25in}

        \caption{Leduc(2, 9)}
    \end{subfigure}
    \hspace*{0.001\linewidth}
    \begin{subfigure}[b]{0.28\linewidth}
        \centering
        \resizebox{\linewidth}{!}{\input{images/plotgoofspiel_4_sbcfr_olo.sh_game_goofspiel_4_cfr_PostSbCFR_a_LinearOpponent_w_Constant_rm_0.1_iter_6.pgf}}
        \vspace*{-0.25in}

        \caption{Goofspiel 4}
    \end{subfigure}
    \\
    \begin{subfigure}[b]{0.28\linewidth}
        \centering
        \resizebox{\linewidth}{!}{\input{images/plotgoofspiel_4_imp_sbcfr_olo.sh_game_goofspiel_4_imp_cfr_PostSbCFR_a_LinearOpponent_w_Constant_rm_0.1_iter_6.pgf}}
        \vspace*{-0.25in}

        \caption{Goofspiel 4(imp.)}
    \end{subfigure}
    \hspace*{0.001\linewidth}
    \begin{subfigure}[b]{0.28\linewidth}
        \centering
        \resizebox{\linewidth}{!}{\input{images/plotFHP2_poker_sbcfr_olo.sh_game_FHP2_poker_cfr_PostSbCFR_a_LinearOpponent_w_Constant_rm_0.0001_iter_6.pgf}}
        \vspace*{-0.25in}

        \caption{FHP(2, 5)}
    \end{subfigure}
    \hspace*{0.001\linewidth}
    \begin{subfigure}[b]{0.28\linewidth}
        \centering
        \resizebox{\linewidth}{!}{\input{images/plotgoofspiel_5_sbcfr_olo.sh_game_goofspiel_5_cfr_PostSbCFR_a_LinearOpponent_w_Constant_rm_0.1_iter_6.pgf}}
        \vspace*{-0.25in}

        \caption{Goofspiel}
    \end{subfigure}
    \\
    \begin{subfigure}[b]{0.28\linewidth}
        \centering
        \resizebox{\linewidth}{!}{\input{images/plotliars_dice_sbcfr_olo.sh_game_liars_dice_cfr_PostSbCFR_a_LinearOpponent_w_Constant_rm_0.1_iter_6.pgf}}
        \vspace*{-0.25in}

        \caption{Liar's dice}
    \end{subfigure}
    \hspace*{0.001\linewidth}
    \begin{subfigure}[b]{0.28\linewidth}
        \centering
        \resizebox{\linewidth}{!}{\input{images/plotbattleship_3_2_3_sbcfr_olo.sh_game_battleship_3_2_3_cfr_PostSbCFR_a_LinearOpponent_w_Constant_rm_0.1_iter_6.pgf}}
        \vspace*{-0.25in}

        \caption{Battleship}
    \end{subfigure}
    \hspace*{0.001\linewidth}
    \hspace*{0.28\linewidth}
    \caption{FD-FTRL(R) with different hyper-parameters in eight games.}
    \label{fig:comp_FD-FTRL(R)_hyper}
\end{figure}

\begin{figure}[htbp]
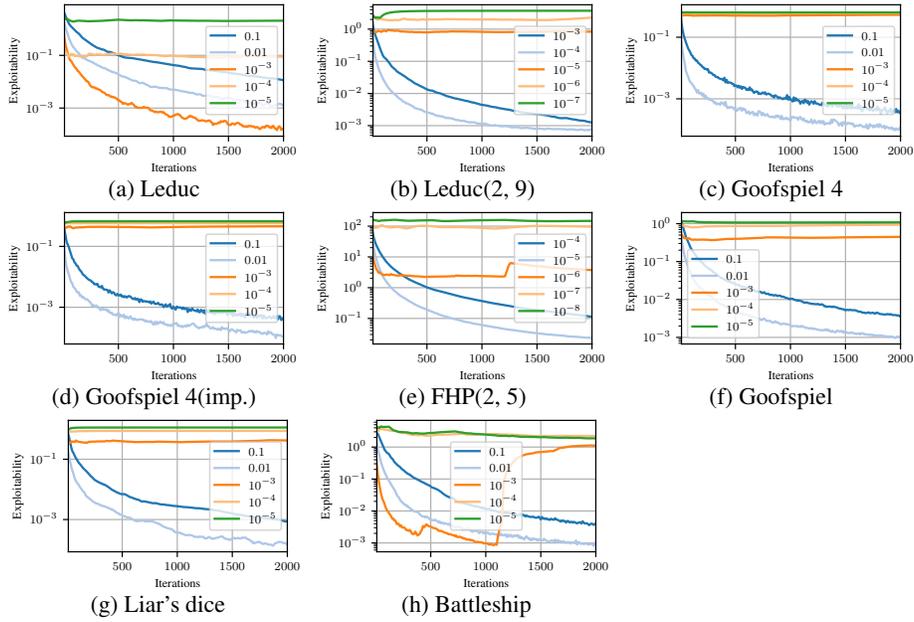

	\centering
    \begin{subfigure}[b]{0.28\linewidth}
        \centering
        \resizebox{\linewidth}{!}{\input{images/plotleduc_poker_sbcfr_olo.sh_game_leduc_poker_cfr_SbCFRPlus_a_LinearOpponent_w_Constant_rm_0.1_iter_10.pgf}}
        \vspace*{-0.25in}

        \caption{Leduc}
    \end{subfigure}
    \hspace*{0.001\linewidth}
    \begin{subfigure}[b]{0.28\linewidth}
        \centering
        \resizebox{\linewidth}{!}{\input{images/plotleduc18_poker_sbcfr_olo.sh_game_leduc18_poker_cfr_SbCFRPlus_a_LinearOpponent_w_Constant_rm_0.001_iter_10.pgf}}
        \vspace*{-0.25in}

        \caption{Leduc(2, 9)}
    \end{subfigure}
    \hspace*{0.001\linewidth}
    \begin{subfigure}[b]{0.28\linewidth}
        \centering
        \resizebox{\linewidth}{!}{\input{images/plotgoofspiel_4_sbcfr_olo.sh_game_goofspiel_4_cfr_SbCFRPlus_a_LinearOpponent_w_Constant_rm_0.1_iter_10.pgf}}
        \vspace*{-0.25in}

        \caption{Goofspiel 4}
    \end{subfigure}
    \\
    \begin{subfigure}[b]{0.28\linewidth}
        \centering
        \resizebox{\linewidth}{!}{\input{images/plotgoofspiel_4_imp_sbcfr_olo.sh_game_goofspiel_4_imp_cfr_SbCFRPlus_a_LinearOpponent_w_Constant_rm_0.1_iter_10.pgf}}
        \vspace*{-0.25in}

        \caption{Goofspiel 4(imp.)}
    \end{subfigure}
    \hspace*{0.001\linewidth}
    \begin{subfigure}[b]{0.28\linewidth}
        \centering
        \resizebox{\linewidth}{!}{\input{images/plotFHP2_poker_sbcfr_olo.sh_game_FHP2_poker_cfr_SbCFRPlus_a_LinearOpponent_w_Constant_rm_0.0001_iter_10.pgf}}
        \vspace*{-0.25in}

        \caption{FHP(2, 5)}
    \end{subfigure}
    \hspace*{0.001\linewidth}
    \begin{subfigure}[b]{0.28\linewidth}
        \centering
        \resizebox{\linewidth}{!}{\input{images/plotgoofspiel_5_sbcfr_olo.sh_game_goofspiel_5_cfr_SbCFRPlus_a_LinearOpponent_w_Constant_rm_0.1_iter_10.pgf}}
        \vspace*{-0.25in}

        \caption{Goofspiel}
    \end{subfigure}
    \\
    \begin{subfigure}[b]{0.28\linewidth}
        \centering
        \resizebox{\linewidth}{!}{\input{images/plotliars_dice_sbcfr_olo.sh_game_liars_dice_cfr_SbCFRPlus_a_LinearOpponent_w_Constant_rm_0.1_iter_10.pgf}}
        \vspace*{-0.25in}

        \caption{Liar's dice}
    \end{subfigure}
    \hspace*{0.001\linewidth}
    \begin{subfigure}[b]{0.28\linewidth}
        \centering
        \resizebox{\linewidth}{!}{\input{images/plotbattleship_3_2_3_sbcfr_olo.sh_game_battleship_3_2_3_cfr_SbCFRPlus_a_LinearOpponent_w_Constant_rm_0.1_iter_10.pgf}}
        \vspace*{-0.25in}

        \caption{Battleship}
    \end{subfigure}
    \hspace*{0.28\linewidth}
    \caption{FD-OMD(R) with different hyper-parameters in eight games.}
    \label{fig:comp_FD-OMD(R)_hyper}
\end{figure}

For each game, We perform a coarse hyper-parameter ($\eta$) tuning for FD-FTRL(R) and FD-OMD(R). For most of the games, we choose $\eta$ in $\{0.1, 0.01, 10^{-3}, 10^{-4}, 10^{-5}\}$. In Figure \ref{fig:comp_FD-FTRL(R)_hyper} (\ref{fig:comp_FD-OMD(R)_hyper}), the results of FD-FTRL(R) (FD-OMD(R)) with different hyper-parameters are given. As we can see, FD-OMD(R) is more sensitive to the hyper-parameter than FD-FTRL(R). In \cite{Recursive_CFR}, the authors have proposed a method for adapting the hyper-parameter in ReCFR, which may be available for adapting the hyper-parameters in FD-FTRL(R) and FD-OMD(R), too.

\end{document}